\documentclass[sigconf,nonacm,screen]{acmart}

\setcopyright{none}
\copyrightyear{2026}
\acmYear{2026}

\usepackage{nicefrac}
\usepackage{pgfplots}
\usepackage{algorithm}
\usepackage{algpseudocode}
\usepackage{listings}
\usepackage{multirow}
\usepackage{array}
\usepackage{xurl}
\pgfplotsset{compat=1.18}

\newcommand{\code}[1]{\path{#1}}

\definecolor{codebg}{RGB}{248,248,248}
\definecolor{codestring}{RGB}{0,92,175}
\definecolor{codekeyword}{RGB}{127,0,85}
\definecolor{codecomment}{RGB}{95,95,95}
\definecolor{codenumber}{RGB}{120,120,120}

\lstdefinestyle{readablecode}{
  basicstyle=\ttfamily\scriptsize,
  backgroundcolor=\color{codebg},
  frame=single,
  framesep=3pt,
  framerule=0.2pt,
  rulecolor=\color{codenumber},
  numbers=none,
  numberstyle=\tiny\color{codenumber},
  numbersep=0pt,
  xleftmargin=0pt,
  framexleftmargin=0pt,
  breaklines=true,
  breakatwhitespace=false,
  columns=fullflexible,
  keepspaces=true,
  showstringspaces=false,
  upquote=true,
  stringstyle=\color{codestring},
  keywordstyle=\color{codekeyword}\bfseries,
  commentstyle=\color{codecomment},
  morestring=[b]",
  morecomment=[l]{//},
  morecomment=[l]{\#},
  morekeywords={true,false,null},
  alsoletter={_},
  aboveskip=0.65em,
  belowskip=0.65em,
}

\lstdefinestyle{worldmodelcode}{
  style=readablecode,
  language=Python,
  basicstyle=\ttfamily\tiny,
  numbers=none,
  numberstyle=\tiny\color{codenumber},
  breaklines=true,
  breakatwhitespace=false,
  columns=fullflexible,
  keepspaces=true,
  showstringspaces=false,
  tabsize=4,
  aboveskip=1em,
  belowskip=1.5em,
}

\newcommand{\methodname}{PatchWorld}
\newcommand{\methodsimple}{\methodname{}-Simple}
\newcommand{\methodresidual}{\methodname{}-Residual}
\newcommand{\loopname}{PatchWorld Repair Loop}
\newcommand{\Score}{\mathrm{Score}}
\newcommand{\emailgrp}[2]{\texttt{\{#1\}@#2}}
\newcommand{\affline}[1]{\hbox to\textwidth{\hss\rule{0pt}{8.5pt}\smash{#1}\hss}\vskip 1.5pt}
\newcommand{\mailline}[1]{\hbox to\textwidth{\hss\rule{0pt}{7pt}\smash{#1}\hss}\vskip 1.5pt}

\begin{document}

\title{PatchWorld: Gradient-Free Optimization of Executable World Models for Agent Environments}

\author{%
  \textbf{Jiaxin Bai}$^{1}$,
  \textbf{Yue Guo}$^{2}$,
  \textbf{Yifei Dong}$^{3}$,
  \textbf{Jiaxuan Xiong}$^{4}$,
  \textbf{Tianshi Zheng}$^{3}$,
  \textbf{Yixia Li}$^{5}$\\[0.15em]
  \textbf{Tianqing Fang}$^{3}$,
  \textbf{Yufei Li}$^{3}$,
  \textbf{Yisen Gao}$^{3}$,
  \textbf{Haoyu Huang}$^{3}$,
  \textbf{Zhongwei Xie}$^{3}$\\[0.15em]
  \textbf{Hong Ting Tsang}$^{3}$,
  \textbf{Zihao Wang}$^{2}$,
  \textbf{Lihui Liu}$^{6}$,
  \textbf{Jeff Z. Pan}$^{7}$,
  \textbf{Yangqiu Song}$^{3}$\\[0.35em]
  {\normalfont\small\offinterlineskip
   \affline{$^{1}$Hong Kong Baptist University \quad
     $^{2}$Independent Researcher \quad
     $^{3}$HKUST}
   \affline{$^{4}$Beijing Institute of Technology \quad
     $^{5}$Southern University of Science and Technology}
   \affline{$^{6}$Wayne State University \quad
     $^{7}$University of Edinburgh}
   \vskip 3pt
   \mailline{\texttt{baijiaxin@hkbu.edu.hk} \quad
     \emailgrp{ireneyueguo}{gmail.com} \quad
     \texttt{xiongjiaxuan@bit.edu.cn}}
   \mailline{\emailgrp{ydongbl, tzhengad, ylivm, ygaodi, hhuangcp, zxiebk}{connect.ust.hk}}
   \mailline{\emailgrp{httsangaj, zwanggc}{connect.ust.hk} \quad
     \emailgrp{tfangaa, yqsong}{cse.ust.hk}}
   \mailline{\texttt{liyixia@me.com} \quad
     \texttt{hw6926@wayne.edu} \quad
     \texttt{j.z.pan@ed.ac.uk}}}
}
\affiliation{%
  \institution{Multiple Institutions}
  \city{Various}
  \country{Various}
}
\renewcommand{\shortauthors}{Bai et al.}
\authorsaddresses{}
\settopmatter{printacmref=false, printccs=false}

\makeatletter
\renewcommand{\@typeset@author@bx}{%
  \bgroup
  \hsize=\textwidth
  \def\and{\par}%
  \normalbaselines
  \global\setbox\author@bx=\vtop{\centering\@authorfont\@currentauthors\par}%
  \box\author@bx
  \gdef\@currentauthors{}%
  \gdef\@currentaffiliation{}%
  \egroup
}
\makeatother

\begin{abstract}
  World models for interactive text agents must typically be learned from
  observation-action trajectories alone.
  Specifically, the environment returns text observations
  after each action, but does not expose a ground-truth latent state nor an
  inspectable transition model.
  A research gap remains in how to induce executable code as a world model in this
  black-box setting for prediction and agent decision making.
  We introduce
  \textbf{PatchWorld}, a gradient-free
  framework that turns offline trajectories into executable Python world models
  through counterexample-guided code repair.
  Instead of predicting the next
  observation with a black-box model, PatchWorld induces symbolic belief-state
  programs whose action updates can be inspected, replayed, and locally patched.
  Across seven AgentGym environments, PatchWorld-Simple achieves the highest
  code-based decision-making score among evaluated methods (76.4\% macro success in live
  one-step lookahead), matching or exceeding LLM-based lookahead while
  invoking no LLM calls inside the world-model prediction module itself.
  We further find that a human-specified residual-memory bias improves surface
  observation fidelity but weakens agent decision-making utility.
  This reveals a tradeoff in
  executable world models, since improving observation fidelity can come at the
  expense of action-discriminative dynamics, and vice versa.
  Code is available at \url{https://github.com/HKBU-KnowComp/PatchWorld}.
\end{abstract}

\maketitle

\section{Introduction}

Scientists advance science by proposing and testing rules and theories that explain observations, predict the effects of interventions, and generalize beyond seen data~\citep{langley1987scientific,schmidt2009distilling,zheng2026newtonbench}.
In artificial agents, world models play an analogous role by learning latent
dynamics for prediction, simulation, agent decision making, and
control~\citep{ha2018worldmodels,hafner2020dreamer,ke2019modeling,lecun2022path},
with recent systems extending these ideas to embodied and generative
environments~\citep{zitkovich2023rt,black2024pi0,bruce2024genie,copet2025cwm}.
For interactive text agents, such models could capture how actions change an
environment and thereby support simulation, diagnosis, and, when useful,
agent decision making before trial-and-error.  
Code-based world models are especially
attractive because they are interpretable, exactly executable,
and locally repairable~\citep{mcdermott1998pddl,wang2024worldcoder}.

\begin{figure*}[t]
  \centering
  \includegraphics[width=\textwidth]{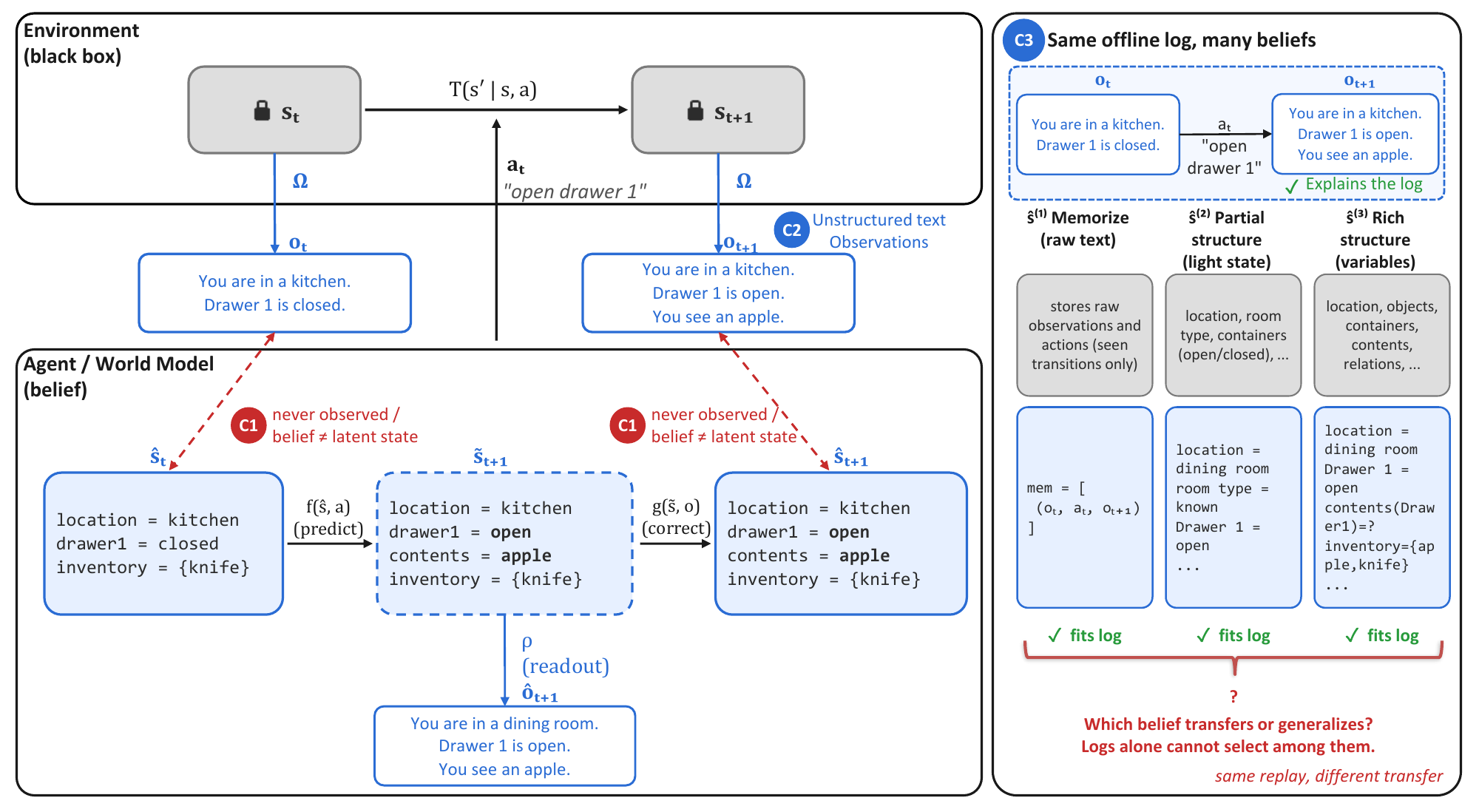}
  \vspace{-0.8cm}
  \caption{Text-agent environments are black-box POMDPs: a latent state
  $s_t$ transitions under action $a_t$ and emits unstructured text
  observations $o_t$, but $s_t$ is never exposed (C1-C2). The agent (and any
  world model) maintains a belief $\hat{s}_t \neq s_t$, updated by predicting
  through actions and correcting from observations. Offline logs alone are
  underdetermined (C3): many beliefs can fit the same
  $(o_t,a_t,o_{t+1})$ while differing in transfer, so replay on held-out
  validation trajectories is needed to probe generalization.}
  \label{fig:teaser}
  \Description{Teaser diagram contrasting a black-box environment latent state
  with an agent belief-state world model, and showing that one offline log can
  be explained by multiple alternative beliefs with different transfer.}
\end{figure*}

Such models have already been explored in Atari-style games and robot control,
where observations are well-structured data or pixel images and dynamics
can often be captured by compact transition
programs~\citep{wang2024worldcoder,piriyakulkij2025poeworld,lehrach2025code,copet2025cwm,black2024pi0}.
They remain far less widely adopted in text-agent environments, including
interfaces, games, and interactive virtual worlds.  Bringing code-based world
models to this setting raises three challenges (Figure~\ref{fig:teaser}).

First, text-agent environments are black-box partially observable Markov
decision processes (POMDPs)~\citep{kaelbling1998planning}: a ground-truth
hidden state may exist, yet the agent never observes it directly and can
recover only what is identifiable from the observation history.  We call the
agent's internal state a \emph{belief}: a representation that approximates
the latent state for prediction and control, and that in general differs from
it ($\hat{s}_t \neq s_t$).
In AlfWorld, a kitchen observation may report that a drawer is closed while
omitting its contents; likewise, Wordle never reveals the target word, and
WebShop hides backend relevance
scores~\citep{cote2018textworld,ALFWorld20,wang2022scienceworld,yao2022webshop,xi2025agentgym,xu-etal-2025-crab}.
Prior text-environment world models often sidestep this issue by predicting the
next observation from recent
history~\citep{li2025wordworldlargelanguage,fang-etal-2025-webevolver}
or by distilling trajectory knowledge into parametric decision-making
aids~\citep{qiao2024wkm}.  These approaches can be useful, but they do not
necessarily maintain the persistent belief needed for multi-step simulation or
agent decision making.

Second, unlike structured pixels or proprioception, text observations are
unstructured natural language.  A useful code-based world model must therefore
maintain a compact belief, apply an action update $f$ to predict the next
belief, render a predicted observation through a readout $\rho$, and correct
the belief with $g$ once the true observation arrives.
In the kitchen example of Figure~\ref{fig:teaser}, opening a closed drawer
may be predicted to reveal an apple even though that object was absent from
the preceding text; the readout then verbalizes the predicted belief, and
correction reconciles it with the observed string.
Encoding both belief dynamics and rendering as executable code is brittle:
the same latent fact may surface in many phrasings, and small wording changes
can break exact replay even when the underlying dynamics are correct.

Third, belief induction from offline trajectories is underdetermined.  Many
different beliefs can explain the same logs: some merely memorize seen
transitions, while others maintain structured variables, such as container
contents or inventory, that support generalization to unseen inputs.
Fitting the training trajectories is therefore insufficient: a hypothesis that
replays perfectly on seen data may still fail on new episodes.
Because training logs alone cannot tell which beliefs will transfer, the
learner needs both an inductive bias toward compact, reusable structure and
an explicit generalization check beyond the induction set.

We propose \textbf{\methodname{}} to address these challenges. \methodname{} is a
gradient-free framework that generates one world model per environment and
treats world-model induction as \emph{generative optimization}.
To address the need for persistent belief, an LLM synthesizes an executable
Python program with an explicit symbolic belief, transition rules, correction
logic, and rendering logic from offline observation-action trajectories only.
To handle unstructured text, the program separates action-discriminative belief
updates from observation readout, so dynamics need not entangle every surface
phrasing.  To constrain underdetermined induction, \methodname{} uses a
program-shaped inductive bias and repairs the hypothesis through
counterexample-guided inductive synthesis
(CEGIS)~\citep{solarlezama2006combinatorial}: trajectories are replayed,
prediction failures become concrete counterexamples, and candidate patches are
accepted only when they improve formal replay fidelity on held-out validation
trajectories rather than on the induction examples alone.
This validation-set replay gate rejects patches that memorize training
transitions without generalizing.  We further evaluate both observation
reconstruction and decision-making utility on a held-out test split, and study
their tradeoff.
The output is not only a next-token predictor, but a reusable and inspectable
executable hypothesis whose belief updates can be run, tested, and locally
repaired.

This dual evaluation matters because world modeling is pulled between two
related goals. Some work emphasizes faithful reconstruction of future
observations~\citep{li2025wordworldlargelanguage,fang-etal-2025-webevolver},
while other work emphasizes downstream decisions through action-contrastive
predictions~\citep{ha2018worldmodels,hafner2020dreamer,lecun2022path,qiao2024wkm}.
Under partial observability, a model can recover surface text while offering
weak action guidance, or miss exact renderings while preserving useful
transition structure (Figure~\ref{fig:pareto}).  \methodname{} therefore treats
text world modeling as a frontier rather than a single optimum.
\methodsimple{} stays on the purely symbolic side of this frontier, while
\methodresidual{} adds a human-specified residual-memory bias for exact,
unambiguous textual state signatures.  This improves reconstruction fidelity
without replacing the executable dynamics used for agent decision making.
More broadly, these results show that LLMs can act as \emph{symbolic
optimizers}, searching, patching, and improving executable symbolic systems
that are competitive with LLM-based world models on both observation
prediction and decision-making utility.

\paragraph{Contributions.}
(1) We formulate three challenges for code-based world models in text-agent
POMDPs: persistent belief under black-box partial observability, unstructured
text observations, and underdetermined belief induction, where finite logs can
support exact replay without revealing compact transferable belief structure,
motivating held-out validation replay as a generalization check
(Section~\ref{sec:feasibility}). (2) We introduce \methodname{}, a
gradient-free framework that uses LLMs as symbolic optimizers to synthesize
and repair executable belief-state programs; its residual-memory variant
treats recurring textual state evidence as a human-specified inductive bias,
not an unconstrained cache (Section~\ref{sec:methodology}). (3) Across seven
AgentGym environments, we find a frontier between observation reconstruction
and decision-making utility. \methodresidual{} attains the highest code-based fidelity,
\methodsimple{} attains the highest code-based live lookahead decision-making score,
and our diagnostics explain where the two goals diverge
(Section~\ref{sec:experiments}).

\begin{figure*}[t]
  \centering
  \includegraphics[width=\textwidth]{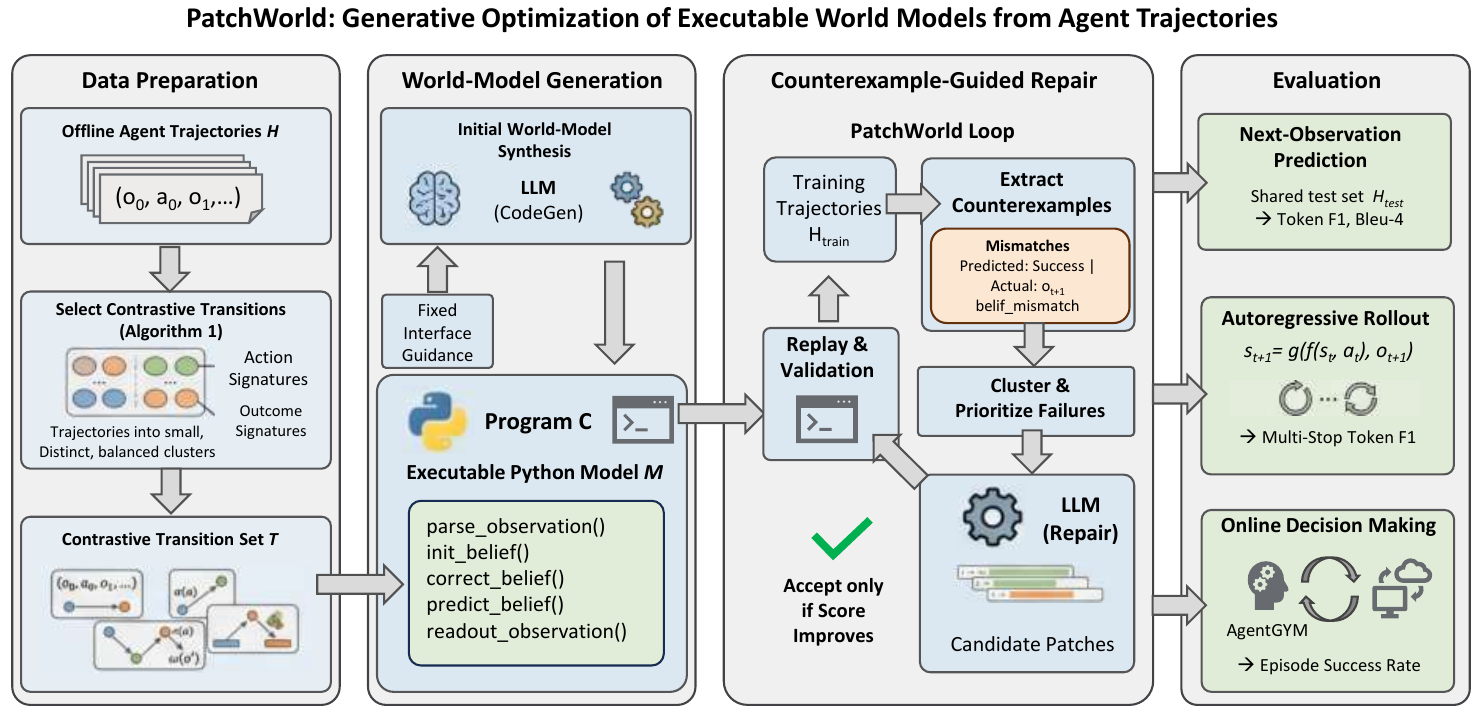}
  \vspace{-0.8cm}
  \caption{Overall framework of PatchWorld for generative optimization of
  executable world models and their subsequent repair and evaluation.}
  \label{fig:patchworld-framework}
  \Description{Framework diagram of PatchWorld induction, counterexample-guided repair, and evaluation.}
\end{figure*}

\section{Related Work}
\label{sec:related_work}

\paragraph{Code-based and symbolic world models.}
A growing line of work represents environment dynamics as executable code or
symbolic rules rather than neural weights
(see~\citet{li2026textworldmodels} for a broader survey of text world models).
In games and control, WorldCoder induces transition programs, GIF-MCTS
searches over code edits, PoE-World composes programmatic experts, and
OneLife learns probabilistic precondition and effect
rules~\citep{wang2024worldcoder,dainese2024generating,piriyakulkij2025poeworld,khan2026one}.
Related efforts target interactive interfaces and planning domains:
Code2World renders GUI next screens as HTML~\citep{zheng2026code2world},
AutoWebWorld synthesizes web finite-state machines~\citep{wu2026autowebworld},
GGP-CWM compiles game rules into OpenSpiel-style Python for
MCTS~\citep{lehrach2025code}, and TheoryCoder and Text2World produce
PDDL or Python action models~\citep{ahmed2025synthesizing,mcdermott1998pddl,text2world2025}.
These systems differ from ours in what is observed during learning and what
the induced artifact must maintain.
POMDP Coder learns probabilistic transition and observation programs, but
requires privileged simulator states $(s_t,a_t,o_t,s_{t+1})$ rather than
text-only logs~\citep{curtis2025pomdp}.
PoE-World and WorldCoder mainly synthesize local transition code and do not
expose a persistent belief interface for partial observability.
\methodname{} instead induces an executable belief-state Python program from
offline text trajectories $(o_t,a_t,o_{t+1})$ only, repairs it with
execution-grounded counterexamples.

\paragraph{Neural, video, and embodied world models.}
A complementary line of work keeps the world model implicit in a neural
module.  In text agents, Word2World fine-tunes a language model to predict
the next observation~\citep{li2025wordworldlargelanguage}; WebEvolver
co-evolves agent and world-model LLMs~\citep{fang-etal-2025-webevolver};
WKM trains a parametric knowledge model for agent decision
making~\citep{qiao2024wkm}; and WALL-E and CoEx maintain rule sets through
online interaction~\citep{zhou2024walle,zhou2025walle2,kim2025coex}.
Beyond text, Genie induces interactive environments from unlabeled
video~\citep{bruce2024genie}, while CWM and $\pi_0$ extend action-conditional
world models to code execution and robot
control~\citep{copet2025cwm,black2024pi0}.
These approaches can improve prediction or control, but the transition logic
stays in weights, prompts, or memories, which limits inspectability and often
requires costly inference-time model calls~\citep{li2026textworldmodels}.
Text-agent observations are also short and templated, so exact tokens matter
more than in pixel domains.
\methodname{} therefore induces an offline, gradient-free program whose belief
can be inspected, executed without LLM calls during lookahead, and repaired
locally when replay fails.

\paragraph{Program synthesis and repair.}
\methodname{}'s induction loop builds on program synthesis from
examples~\citep{gulwani2017program,ellis2021dreamcoder,wang2024hypothesis,wei2025codearc}
and instantiates counterexample-guided inductive
synthesis~\citep{solarlezama2006combinatorial} for sequential POMDP
dynamics. In this setting, the specification is a set of replayable
trajectories, the verifier is an executable validator that reports typed
failures, and the synthesizer is an LLM prompted with structured diagnostics,
making the LLM a gradient-free symbolic optimizer rather than the transition
model itself.
Related logical-hypothesis work studies constraint-based
reasoning in knowledge graphs~\citep{bai2023complex,bai2024advancing}, whereas
our hypotheses are executable transition programs.
Compared with agent reflection and decision-making adaptation
methods~\citep{shinn2023reflexion,wang2023voyager,prasad-etal-2024-adapt,yuan2025agentr,kim-etal-2025-reflact}
and LLM code self-repair from execution
feedback~\citep{quoc2024empirical,kamoi2024when,jiang2024ledex,islam-etal-2025-codesim},
\methodname{} addresses a failure mode of hallucinated or weakly checked
code~\citep{agarwal2024codemirage,councilman2025towards}: it accepts a patch
only when the patch improves formal replay score on the full validation set,
preventing the regression problem that plagues unconstrained self-correction.

\section{Task Formulation}
\label{sec:task_formulation}

We model each environment as a POMDP
$(\mathcal{S}, \mathcal{A}, \mathcal{T}, \Omega, \mathcal{O})$ with latent
states $s_t\in\mathcal{S}$, text actions $a_t\in\mathcal{A}$, observations
$o_t\in\mathcal{O}$, transition kernel
$\mathcal{T}(s_{t+1}\!\mid\!s_t,a_t)$, and emission $\Omega(o_t\!\mid\!s_t)$.
The learner sees neither $s_t$ nor $\mathcal{T}$, only offline trajectories
$\mathcal{H}=\{(o_0,a_0,\ldots,a_{T-1},o_T)\}$ and an optional
description~$E$.

\paragraph{Goal.}
Induce executable Python $c$ that approximates dynamics through a symbolic
belief $\hat{s}_t \in \hat{\mathcal{S}}$ and a predict and correct cycle. The
program implements $f_c$ (\emph{predict}), $\rho_c$ (\emph{readout}), and $g_c$
(\emph{correct}). Let $\bot$ denote the blank belief returned by
\code{init_belief}; \code{correct_belief} then populates it from the
initial observation:
\begin{align}
\hat{s}_0 &= g_c(\bot,\, o_0), \label{eq:belief_init}\\
\tilde{s}_{t+1} &= f_c(\hat{s}_t,\, a_t), \label{eq:belief_predict}\\
\hat{o}_{t+1} &= \rho_c(\tilde{s}_{t+1},\, a_t), \label{eq:belief_render}\\
\hat{s}_{t+1} &= g_c(\tilde{s}_{t+1},\, o_{t+1}). \label{eq:belief_correct}
\end{align}
The induction objective is to minimize replay loss
\begin{equation}
\min_{c}\;
\mathbb{E}_{(o_t,a_t,o_{t+1})\sim\mathcal{H}}
\bigl[\ell(\hat{o}_{t+1},\, o_{t+1})\bigr],
\label{eq:induction_objective}
\end{equation}
where $\ell$ is a text loss over predicted and observed next observations and
$\hat{s}_t$ comes from running
\eqref{eq:belief_init}-\eqref{eq:belief_correct} along the prefix. Since $c$
is discrete, we optimize by search
(Section~\ref{sec:methodology}) using typed replay validation and a
severity-weighted error score (Eq.~\eqref{eq:replay-score}) rather than
differentiating $\ell$ directly; the output is an inspectable program whose
beliefs, rules, and render logic can be locally patched.

The program $c$ may also include a constrained residual readout memory for
high-confidence surface details, while preserving the predict and correct
interface. These constraints are an inductive bias, analogous to weight decay,
that steers search toward compact explanations
(Section~\ref{sec:residual-memory}).

\paragraph{Deterministic belief approximation.}
The learning target is a deterministic executable belief update under partial
observability: given history and action, $c$ produces the next belief and
observation string.
This matches the needs of text-agent prediction and one-step lookahead, where
the agent must maintain an actionable summary of what has been observed
(e.g., a Wordle candidate set) and step it forward under interventions.
Stochastic kernels over latent simulator states are a compatible extension of
the same interface and are left for future work.

\paragraph{Feasibility and identification.}
\label{sec:feasibility}
Offline induction from $\mathcal{H}$ is a well-posed search problem with a
standard identification structure~\citep{kaelbling1998planning}.
Many programs can replay the logged transitions, ranging from prefix-action
lookup tables to compact rule systems, so replay fitness alone under-specifies
which hypothesis will transfer.
\methodname{} therefore treats inductive bias as part of the method: an LLM
code prior, contrastive counterexample-guided repair that enforces consistency
across diverse transition patterns, and explicit constraints on residual
memory.
Together these preferences select structured, reusable programs over
rote memorization of $\mathcal{H}$.

\paragraph{Fidelity and decision utility.}
\label{sec:fidelity-utility}
Eq.~\eqref{eq:induction_objective} measures textual reconstruction of
$o_{t+1}$.
Agent decision making instead depends on whether predicted futures separate
good actions from bad ones.
Let $\Delta_c(s,a,a')$ denote the action-contrastive difference between the
readouts $\rho_c(f_c(s,a),a)$ and $\rho_c(f_c(s,a'),a')$ that the selector
uses for ranking.
Reconstruction and $\Delta_c$ are related but distinct objectives: surface
factors that are shared across actions (template phrasing, identifiers, list
orderings) can change $\ell$ while leaving action rankings unchanged, and
conversely~\citep{grimm2020value}.
Accordingly, we evaluate both observation fidelity and live lookahead success
and characterize their Pareto frontier in
\S\ref{sec:finding-planning} and Appendix~\ref{app:contrast-diagnostics}.

\begin{table}[t]
\caption{Environment regimes vs.\ intro challenges (C1 belief, C2 text, C3 underdetermination).}
\vspace{-0.3cm}
\label{tab:env-regimes}
\small\centering
\setlength{\tabcolsep}{3pt}
\begin{tabular*}{\columnwidth}{@{\extracolsep{\fill}}lll@{}}
\toprule
\textbf{Regime} & \textbf{Envs.} & \textbf{Stress} \\
\midrule
Belief collapses w/ exploration & Maze, BabyAI, TextCraft & C1 \\
Belief stays underdetermined & Wordle, WebShop & C1+C3 \\
Dynamics OK, rendering brittle & AlfWorld, SciWorld & C2 \\
\bottomrule
\end{tabular*}
\end{table}

\begin{figure}[t]
  \centering
  \begin{lstlisting}[style=readablecode,language=Python,numbers=none]
  class BaseWorldModel:
  
      # Parse text into facts.
      def parse_observation(self, obs: str) -> dict: ...
  
      # Initialize blank latent belief.
      def init_belief(self) -> State: ...
  
      # Populate / update belief from text.
      def correct_belief(self, state, obs: str) -> State: ...
  
      # Apply action dynamics.
      def predict_belief(self, state, action: str) -> State: ...
  
      # Render next observation.
      def readout_observation(self, state, action: str) -> str: ...
  \end{lstlisting}
  \vspace{-0.3cm}
  \caption{Core executable interface implemented by each induced program,
  separating parsing, belief correction, dynamics, and readout.}
  \label{fig:model-interface}
\Description{Python class skeleton listing BaseWorldModel interface methods.}
  \end{figure}

\section{Methodology}
\label{sec:methodology}

\methodname{} induces a standalone Python world model for each environment by
execution-grounded search over programs
(Figure~\ref{fig:patchworld-framework}).
Given offline trajectories, the method selects contrastive evidence, prompts an
LLM to synthesize a complete \code{BaseWorldModel}, replays the candidate to
expose counterexamples, and accepts a repair only when it improves validation
replay.
The resulting program maintains an explicit belief state, supports agent
decision making, and predicts next observations without additional LLM calls at
inference time.
Unless otherwise stated, induction uses the training split, repair uses the
validation split when available (otherwise training), and evaluation is
reserved for the held-out test split.

\subsection{Executable World-Model Interface}

Each induced module implements the predict and correct interface defined in
Section~\ref{sec:task_formulation} and summarized in
Figure~\ref{fig:model-interface}.
During replay, belief is corrected with the logged observation before the next
action is applied; during rollout, the model is conditioned on its own
predicted observations so that compounding errors become visible to the
validator.
The synthesis prompt specifies the target environment, but does not prescribe a
fixed belief schema: the LLM proposes a state representation suited to that
environment's observations and dynamics.

\subsection{Evidence Selection via Contrastive Mining}

Because the synthesis prompt can include only a small subset of
$\mathcal{H}$, \methodname{} mines a compact contrastive set that covers
diverse action and outcome patterns
(Algorithm~\ref{alg:contrastive}).
Each transition $\tau=(o,a,o')$ is indexed by two signatures.
The action signature $\alpha(a)$ replaces instance-specific arguments by
types (for example,
\texttt{take mug 1 from shelf 2} becomes
\texttt{take OBJ from RECEPTACLE}).
The outcome signature $\omega(o')$ labels the next observation as a
successful state change, a no-op or invalid response, a terminal outcome, or
an informational reply.
We then retain, for each action type, both success-like and failure-like
continuations, so the induced program must learn conditional rules rather than
a single memorized string.
Algorithm~\ref{alg:contrastive} stores at most $k$ transitions per bucket
$(\alpha,\omega)$ and samples across buckets in round-robin order until at
most $m$ examples are collected.
Replay validation continues to use the full trajectory set; only the LLM
prompt is restricted to the mined subset.

\begin{algorithm}[t]
  \caption{Contrastive transition selection}
  \label{alg:contrastive}
  \begin{algorithmic}[1]
  \Require Induction trajectories $\mathcal{H}_{\text{ind}}$, caps $k,m$
  \State initialize buckets $B_{\alpha,\omega}$, each holding $\leq k$ transitions
  \State $T \gets \emptyset$; let $\mathcal{A}$ be signatures with non-empty buckets
  \While{$|T|<m$ and buckets remain}
      \State round-robin pick $\alpha\in\mathcal{A}$; pop one $\tau$ from any non-empty $B_{\alpha,\omega}$
      \State $T \gets T \cup \{\tau\}$
  \EndWhile
  \Ensure Contrastive multiset $T$
  \end{algorithmic}
\end{algorithm}

\subsection{Counterexample-Guided Repair}
\label{sec:induction}

Induction begins from a synthesis prompt that includes the contrastive set
$T$, an environment description, the \code{BaseWorldModel} interface contract,
and instructions to emit a complete executable module.
Thereafter, \methodname{} searches over programs by proposing candidates with
an LLM and accepting updates only when replay validation improves
(Algorithm~\ref{alg:induce}).

\paragraph{Typed replay validation.}
Eq.~\eqref{eq:induction_objective} defines the target as a text loss $\ell$
between $\hat{o}_{t+1}$ and $o_{t+1}$.
Because $c$ is discrete, search does not differentiate $\ell$; instead the
validator operationalizes replay disagreement through typed failures on a
replay set $B$.
Given a candidate $c$, it executes the predict and correct cycle in
\eqref{eq:belief_init}-\eqref{eq:belief_correct} on each logged trajectory
and forms
$\hat{o}^{(c)}_{t+1}=\rho_c(f_c(\hat{s}_t,a_t),a_t)$.
A failed check yields a typed counterexample (load error, parser exception,
belief-update error, task-specific state mismatch, or readout mismatch).
Let $\mathcal{E}(c;B)$ denote this set.
Held-out prediction evaluation separately reports continuous instantiations
of $\ell$ (Token~F1 and BLEU-4); episode success is reserved for the
decision-making experiments in Section~\ref{sec:experiments}.

\paragraph{Repair prompt construction.}
Two operators turn $\mathcal{E}(c;B)$ into an editable diagnosis.
\textsc{Diagnose} clusters related failures and summarizes the dominant pattern
in natural language (for example, that \texttt{move} predictions drop
\code{walls_visible} when a move is blocked).
\textsc{Prioritize} ranks individual failures so that execution and parser
errors precede milder semantic mismatches.
Each repair prompt then presents the current program, the diagnosis, the
top-16 counterexamples (input, expected output, and actual output), and the
fixed interface guidance; the LLM returns a full replacement module.

\paragraph{Acceptance and stopping.}
As a searchable surrogate for $\ell$ on $B$, candidate quality is measured by
the severity-weighted replay score
\begin{equation}
\Score(c;B)=\sum_{e\in\mathcal{E}(c;B)}\bigl(1+\tfrac{1}{4}\,\sigma(e)\bigr),
\label{eq:replay-score}
\end{equation}
where $\sigma(e)$ is a fixed rank over error types and non-executable programs
incur a large penalty.
Lower scores correspond to fewer or less severe disagreements with the logged
observations.
In each round we decode $C$ independent candidates and accept at most one,
requiring a strict improvement in $\Score(c;B)$ and using
$|\mathcal{E}(c;B)|$ as a tie-breaker.
This gate rejects patches that repair only the displayed examples while
regressing on the broader replay set $B$.
Search terminates when $\mathcal{E}=\emptyset$, the round budget is exhausted,
or no candidate improves the score; the accepted program is written out as a
Python module.
At inference time, next-observation prediction uses only
\code{predict_belief} followed by \code{readout_observation} (no neural
renderer); parser outputs $\delta_t=\pi_c(o_t)$ may be inspected during replay
for diagnostics.
Figure~\ref{fig:qualitative-repair} illustrates a Maze repair in which a
concrete counterexample leads to a localized patch that tightens terminal
parsing while preserving wall information on blocked moves.

\begin{figure}[t]

\Description{Code listing of a qualitative counterexample and accepted patch for Maze.}

\begin{lstlisting}[style=readablecode]
# Counterexample
observation:  "The goal is at position 8, 6. ... position 1, 2.
               There are walls above you, below you."
action:       "move right"
expected:     "... position 1, 3. There is a wall above you."
before patch: "Success" or "... There are no walls around you."

# Accepted patch
- if "Success" in obs_text:
+ if obs_text.strip() == "Success":
      result["status"] = "success"
  if move_blocked:
-     pass
+     new_belief["local_walls"] = belief["local_walls"].copy()
\end{lstlisting}
\vspace{-0.3cm}
\caption{\textbf{Qualitative PatchWorld repair example.}
The validator turns a failed Maze transition into a localized code change;
see Appendix~\ref{app:qualitative-repair}.}
\label{fig:qualitative-repair}

\end{figure}

\begin{algorithm}[t]
\caption{\loopname{}: counterexample-guided patch search}
\label{alg:induce}
\begin{algorithmic}[1]
\Require Train $\mathcal{H}_{\text{train}}$, optional val $\mathcal{H}_{\text{val}}$, caps $k,m$, budget $R$, candidates $C$, beam $b$
\State $\mathcal{H}_{\text{replay}} \gets \mathcal{H}_{\text{val}}$ if available else $\mathcal{H}_{\text{train}}$
\State $T \gets \textsc{SelectContrastive}(\mathcal{H}_{\text{train}},k,m)$
\State $G \gets$ interface and transition guidance
\State $c \gets \text{LLM}_{\text{code}}(G,T)$
\State $\mathcal{E} \gets \textsc{Validate}(c,\mathcal{H}_{\text{replay}})$
\State $q \gets \Score(c;\mathcal{H}_{\text{replay}})$
\For{$r = 1$ \textbf{to} $R$}
    \If{$\mathcal{E}=\emptyset$} \textbf{break} \EndIf
    \State $D,\bar{\mathcal{E}} \gets \textsc{Diagnose}(\mathcal{E}),\,\textsc{Prioritize}(\mathcal{E})$
    \For{$j = 1$ \textbf{to} $C$}
        \State $\tilde{c}_j \gets \text{LLM}_{\text{repair}}(c,\bar{\mathcal{E}}_{1:16},D,G;\operatorname{seed}=j)$
        \State $\tilde{\mathcal{E}}_j \gets \textsc{Validate}(\tilde{c}_j,\mathcal{H}_{\text{replay}})$
        \State $\tilde{q}_j \gets \Score(\tilde{c}_j;\mathcal{H}_{\text{replay}})$
    \EndFor
    \State retain top $b$ by $(\tilde{q}_j,|\tilde{\mathcal{E}}_j|)$
    \If{some retained candidate has $\tilde{q}_j<q$ or $(\tilde{q}_j=q \textbf{ and } |\tilde{\mathcal{E}}_j|<|\mathcal{E}|)$}
        \State $c,\mathcal{E},q \gets$ best improving candidate
    \Else
        \State \textbf{break}
    \EndIf
\EndFor
\Ensure Final program $c$
\end{algorithmic}
\end{algorithm}

\subsection{Inductive Bias for Fidelity}
\label{sec:residual-memory}

\methodsimple{} uses only the induced symbolic program.
This is often enough for decision making, but compact rules can miss exact
surface text such as object IDs or long product listings.
\methodresidual{} therefore adds a small train-only lookup table for those
recurring strings, while belief updates still go through
\code{predict_belief} and \code{correct_belief}.
Each training pair (observation, action) is stored under a normalized key
(lower-cased, whitespace collapsed; object IDs kept), together with its
majority next observation, but only when that next observation is unique for
the key.
At test time, a key hit returns the stored string; otherwise the symbolic
readout is used.
Because the table is built from training data only and rejects conflicting
outcomes, it cannot memorize validation or test transitions, and it avoids
aliases such as identical Wordle screens with different hidden targets.
Details are in Appendix~\ref{app:residual-coverage}.

\begin{table*}[t]
  \caption{One-step next-observation prediction on the held-out test set, with
  Token F1 (F1) and BLEU-4 (BL); Avg.\ is the unweighted mean. ``Infer.\ LLM?''
  indicates whether test-time prediction needs LLM calls after fitting. Bold
  marks the best code-based score.}
  \vspace{-0.3cm}
  \label{tab:rq1-onestep}
  \scriptsize
  \centering
  \setlength{\tabcolsep}{2pt}
  \resizebox{\textwidth}{!}{%
  \begin{tabular}{@{}llll *{8}{cc}@{}}
  \toprule
  Method & Backbone & LLM Usage & Infer. LLM?
    & \multicolumn{2}{c}{AlfWorld}
    & \multicolumn{2}{c}{BabyAI}
    & \multicolumn{2}{c}{Maze}
    & \multicolumn{2}{c}{SciWorld}
    & \multicolumn{2}{c}{TextCraft}
    & \multicolumn{2}{c}{WebShop}
    & \multicolumn{2}{c}{Wordle}
    & \multicolumn{2}{c}{Avg.} \\
  \cmidrule(lr){5-6}\cmidrule(lr){7-8}\cmidrule(lr){9-10}\cmidrule(lr){11-12}
  \cmidrule(lr){13-14}\cmidrule(lr){15-16}\cmidrule(lr){17-18}\cmidrule(lr){19-20}
    & & & & F1 & BL & F1 & BL & F1 & BL & F1 & BL & F1 & BL & F1 & BL & F1 & BL & F1 & BL \\
  \midrule
  \multicolumn{20}{@{}l}{\textit{LLM-based Neural World Models}} \\
  \midrule
  Word2World~\citep{li2025wordworldlargelanguage} & Qwen3.5-4B & SFT & Yes
    & 0.89 & 0.66 & 0.93 & 0.75 & 0.97 & 0.89 & 0.96 & 0.95 & 0.94 & 0.68 & 0.63 & 0.42 & 0.60 & 0.37 & 0.85 & 0.67 \\
  \cmidrule(l){2-20}
  \multirow{3}{*}{LLM-Direct} & Qwen3-Coder-480B & ICL & Yes
    & 0.53 & 0.35 & 0.73 & 0.44 & 0.83 & 0.75 & 0.48 & 0.30 & 0.76 & 0.51 & 0.56 & 0.34 & 0.60 & 0.21 & 0.64 & 0.41 \\
  & Mimo-v2.5 & ICL & Yes
    & 0.55 & 0.33 & 0.81 & 0.54 & 0.83 & 0.75 & 0.45 & 0.25 & 0.76 & 0.50 & 0.52 & 0.25 & 0.51 & 0.24 & 0.63 & 0.41 \\
  & DeepSeek-V4-Flash & ICL & Yes
    & 0.55 & 0.34 & 0.81 & 0.56 & 0.83 & 0.73 & 0.52 & 0.34 & 0.78 & 0.51 & 0.58 & 0.36 & 0.53 & 0.19 & 0.66 & 0.43 \\
  \midrule
  \multicolumn{20}{@{}l}{\textit{Program-based World Models (No LLMs at Inference)}} \\
  \midrule
  POMDP Coder~\citep{curtis2025pomdp} & Qwen3-Coder-480B & CodeGen & No
    & 0.67 & 0.42 & 0.59 & 0.35 & 0.85 & 0.70 & 0.40 & 0.32 & 0.73 & 0.48 & 0.24 & 0.05 & 0.00 & 0.00 & 0.50 & 0.33 \\
  \cmidrule(l){2-20}
  GGP-CWM~\citep{lehrach2025code} & Qwen3-Coder-480B & CodeGen & No
    & 0.26 & 0.08 & 0.42 & 0.14 & 0.71 & 0.51 & 0.00 & 0.00 & 0.24 & 0.12 & 0.36 & 0.12 & 0.00 & 0.00 & 0.29 & 0.14 \\
  \cmidrule(l){2-20}
  \multirow{3}{*}{WorldCoder~\citep{wang2024worldcoder}} & Qwen3-Coder-480B & CodeGen & No
    & 0.63 & 0.42 & 0.78 & 0.61 & 0.83 & 0.76 & 0.40 & 0.36 & 0.88 & 0.61 & 0.58 & 0.45 & 0.31 & 0.11 & 0.63 & 0.48 \\
   & Mimo-v2.5 & CodeGen & No
    & 0.59 & 0.35 & 0.77 & 0.61 & 0.88 & 0.78 & 0.37 & 0.33 & 0.88 & 0.61 & 0.53 & 0.37 & 0.35 & 0.13 & 0.63 & 0.46 \\
   & DeepSeek-V4-Flash & CodeGen & No
    & 0.40 & 0.21 & 0.70 & 0.54 & 0.73 & 0.61 & 0.41 & 0.33 & 0.60 & 0.35 & 0.58 & 0.45 & 0.55 & 0.22 & 0.57 & 0.39 \\
  \cmidrule(l){2-20}
  \multirow{3}{*}{PoE-World~\citep{piriyakulkij2025poeworld}} & Qwen3-Coder-480B & CodeGen & No
    & 0.62 & 0.40 & 0.78 & \textbf{0.62} & 0.83 & 0.77 & 0.39 & 0.34 & 0.78 & 0.55 & 0.58 & 0.44 & 0.43 & 0.17 & 0.63 & 0.47 \\
   & Mimo-v2.5 & CodeGen & No
    & 0.62 & 0.40 & 0.78 & \textbf{0.62} & 0.86 & 0.78 & 0.41 & 0.36 & 0.88 & 0.61 & \textbf{0.60} & \textbf{0.47} & 0.40 & 0.16 & 0.65 & 0.49 \\
   & DeepSeek-V4-Flash & CodeGen & No
    & 0.37 & 0.14 & 0.73 & 0.54 & 0.76 & 0.61 & 0.21 & 0.21 & 0.50 & 0.41 & 0.47 & 0.33 & 0.55 & 0.22 & 0.51 & 0.35 \\
  \cmidrule(l){2-20}
  \multirow{3}{*}{\methodsimple{} (ours)} & Qwen3-Coder-480B & CodeGen & No
    & 0.36 & 0.10 & \textbf{0.85} & 0.58 & 0.80 & 0.69 & 0.57 & 0.39 & 0.28 & 0.10 & 0.53 & 0.30 & 0.58 & 0.20 & 0.57 & 0.34 \\
   & Mimo-v2.5 & CodeGen & No
    & 0.73 & 0.47 & 0.41 & 0.19 & 0.88 & 0.75 & 0.48 & 0.30 & 0.93 & 0.67 & 0.28 & 0.12 & 0.47 & 0.16 & 0.60 & 0.38 \\
   & DeepSeek-V4-Flash & CodeGen & No
    & 0.48 & 0.21 & 0.77 & 0.51 & 0.90 & 0.80 & 0.22 & 0.04 & 0.84 & 0.61 & 0.28 & 0.10 & 0.69 & 0.40 & 0.60 & 0.38 \\
  \cmidrule(l){2-20}
  \multirow{3}{*}{\methodresidual{} (ours)} & Qwen3-Coder-480B & CodeGen & No
    & 0.70 & 0.47 & 0.69 & 0.47 & 0.97 & 0.91 & 0.56 & 0.48 & 0.71 & 0.48 & 0.50 & 0.26 & \textbf{0.72} & \textbf{0.44} & 0.69 & \textbf{0.50} \\
   & Mimo-v2.5 & CodeGen & No
    & \textbf{0.77} & \textbf{0.50} & 0.49 & 0.28 & 0.87 & 0.82 & \textbf{0.69} & \textbf{0.56} & \textbf{0.95} & \textbf{0.68} & 0.50 & 0.26 & 0.63 & 0.35 & \textbf{0.70} & 0.49 \\
   & DeepSeek-V4-Flash & CodeGen & No
    & 0.57 & 0.29 & 0.80 & 0.56 & \textbf{0.98} & \textbf{0.93} & 0.56 & 0.46 & 0.91 & 0.66 & 0.50 & 0.26 & 0.60 & 0.34 & \textbf{0.70} & \textbf{0.50} \\
  \bottomrule
  \end{tabular}}
\vspace{-0.1cm}
  \end{table*}

\section{Experiments}
\label{sec:experiments}

We evaluate \methodname{} on observation fidelity and decision-making utility. One-step
and rollout prediction test future-observation match, while live one-step
lookahead tests action choice. The two evaluations favor different variants.
\methodresidual{} is the best code-based predictor, while \methodsimple{} is
the best code-based decision-making agent.

\subsection{Setup}
\label{sec:setup}

\paragraph{Environments and data.}
We evaluate on seven AgentGym~\citep{xi2025agentgym} environments: Maze~\citep{abdulhai2025lmrl},
BabyAI~\citep{chevalier-boisvert2018babyai}, TextCraft~\citep{prasad-etal-2024-adapt},
Wordle~\citep{abdulhai2025lmrl}, WebShop~\citep{yao2022webshop},
AlfWorld~\citep{ALFWorld20}, and SciWorld~\citep{wang2022scienceworld}.
Together they cover navigation, partial observability, crafting, shopping, and
long-horizon household tasks.
Offline trajectories are collected with a strong coding LLM under a ReAct
policy\footnote{Qwen3-Coder-480B-A35B-Instruct:
\url{https://huggingface.co/Qwen/Qwen3-Coder-480B-A35B-Instruct}.}
and split 60/20/20 by instance ID, so all rollouts from one instance stay in
the same split.
Training data are used for world-model induction (and Word2World fine-tuning);
validation data for repair and hyperparameter choices; and test data only for
final reporting.
For each environment and backbone we induce one program with fixed repair
settings; any residual-memory index is built from training transitions only. Details are in Appendix~\ref{app:dataset-stats}.

\paragraph{Baselines.}
\textbf{LLM-Direct} predicts with $k{=}3$ in-context transitions.
\textbf{Word2World}~\citep{li2025wordworldlargelanguage} is an SFT
implicit predictor.
On the code side, \textbf{WorldCoder}~\citep{wang2024worldcoder} and
\textbf{PoE-World}~\citep{piriyakulkij2025poeworld} induce transition
programs;
\textbf{POMDP Coder}~\citep{curtis2025pomdp} and
\textbf{GGP-CWM}~\citep{lehrach2025code} are recent executable
world-model baselines that we adapt to text-only AgentGym supervision;
adaptation details appear in Appendix~\ref{app:pomdp-code-baselines}.
\textbf{ReAct}~\citep{yao2023react} is used only in agent decision making.
In live decision making, Word2World remains a neural next-observation
predictor at inference time, whereas the code-based methods execute induced
programs.
All code-induction methods share the same splits, backbone per row, and
per-environment LLM budget caps; \methodname{} uses contrastive caps
$k{=}5,m{=}60$.
We report Token F1 and BLEU-4 for prediction, and episode success for live
decision making.
All LLM calls use Qwen3-Coder-480B unless noted.
Word2World uses Qwen3.5-4B; selected code-induction rows also report Mimo-v2.5
and DeepSeek-V4-Flash, with model cards in Table~\ref{tab:rq1-onestep}.
\subsection{One-Step Observation Prediction}
\label{sec:finding-onestep}

On each held-out transition, every method predicts $\hat{o}_{t+1}$ from
$(o_t,a_t)$ with no per-instance adaptation; \methodname{} does so through
belief update then readout, as reported in Table~\ref{tab:rq1-onestep}.

\paragraph{Analysis.}
\textbf{\methodresidual{} attains the strongest program-based one-step
scores: 0.69 Token~F1 / 0.50 BLEU-4 with Qwen3-Coder-480B, and up to 0.70
Token~F1 with Mimo-v2.5 or DeepSeek-V4-Flash, exceeding matched WorldCoder
and PoE-World rows by roughly 6--7 points.}
Among executable methods, \methodresidual{} outperforms WorldCoder and
PoE-World, which in turn outperform \methodsimple{}.
\methodsimple{} already indicates that belief tracking is largely effective:
under Qwen3-Coder-480B it reaches 0.57 macro Token~F1, and the validator
analysis in Appendix~\ref{app:error-analysis} attributes most errors to
readout mismatches rather than incorrect state updates.
The residual deficit is therefore primarily surface realization. On
template-heavy environments such as AlfWorld and TextCraft, where exact
string match matters, \methodsimple{} trails WorldCoder and PoE-World, both
at 0.63 macro Token~F1 under the same backbone.
As shown in Table~\ref{tab:rq1-onestep}, constrained train-only residual
memory narrows this gap to 0.69 Token~F1 while remaining an executable
program rather than an end-to-end neural predictor.
Exact train keys cover only 33\% of held-out transitions on average, and a
retrieval-only predictor scores 0.32 when misses are scored as zero;
Appendices~\ref{app:residual-coverage} and~\ref{app:seed-variance} report
coverage and show that the residual margin exceeds induction variance.

\subsection{Multi-Step Rollout}
\label{sec:finding-rollout}

We start from ground-truth $o_0$ and roll out with ground-truth actions but
the model's own predicted observations,
\begin{equation*}
\hat{o}_{t+1}=\rho_c\bigl(f_c(\hat{s}_t,a_t),a_t\bigr),\qquad
\hat{s}_{t+1}\leftarrow g_c\bigl(f_c(\hat{s}_t,a_t),\hat{o}_{t+1}\bigr),
\end{equation*}
for up to five steps; Table~\ref{tab:rq2-macro-rollout} reports the
macro averages, with per-environment details in
Appendix~\ref{app:rq2-full-rollout}.
Scores at $t{=}1$ use episode-level filtering and are not directly comparable
to Section~\ref{sec:finding-onestep}.

\begin{table}[t]
\caption{Average rollout Token F1, unweighted over the seven
environments. Bold marks the best code-based score. Per-environment results
are in Appendix~\ref{app:rq2-full-rollout}.}
\vspace{-0.3cm}
\label{tab:rq2-macro-rollout}
\centering
\setlength{\tabcolsep}{4pt}
\begin{tabular}{@{}lcccc@{}}
\toprule
\textbf{Method} & \textbf{$t{=}1$} & \textbf{$t{=}2$} & \textbf{$t{=}3$} & \textbf{$t{=}5$} \\
\midrule
LLM-Direct & 0.64 & 0.63 & 0.60 & 0.59 \\
\cmidrule(l){1-5}
WorldCoder & 0.63 & 0.60 & 0.58 & 0.57 \\
PoE-World  & 0.63 & 0.57 & 0.53 & 0.51 \\
\methodsimple{} (ours)   & 0.56 & 0.50 & 0.43 & 0.40 \\
\methodresidual{} (ours) & \textbf{0.69} & \textbf{0.64} & \textbf{0.60} & \textbf{0.58} \\
\bottomrule
\end{tabular}
\vspace{-0.3cm}
\end{table}

\paragraph{Analysis.}
\textbf{\methodresidual{} attains the strongest code-based rollout at every
horizon, with 0.58 Token~F1 at $t{=}5$ against 0.57 for WorldCoder and 0.51
for PoE-World.}
Multi-step rollout stresses partial observability over time: each prediction
is fed back as the next observation, so small wording errors can compound
unless the model re-grounds them in a persistent belief.
\methodname{} does this explicitly by correcting belief from its own readout
before the next action.
As reported in Table~\ref{tab:rq2-macro-rollout}, \methodresidual{} remains
strongest from $t{=}1$ through $t{=}5$; its five-step score nearly matches
WorldCoder and substantially exceeds PoE-World.
Even \methodsimple{}, despite weaker surface text of 0.56 at $t{=}1$, remains
usable at longer horizons with 0.40 at $t{=}5$: the structured belief still
carries enough latent state to limit cascade failures after the first
mismatch.
Executable belief state therefore supports not only next-token fidelity, but
also stability when observations must be imagined over several steps.

\subsection{Live Decision Making}
\label{sec:finding-planning}

\begin{figure}[t]
\centering
\includegraphics[width=\columnwidth]{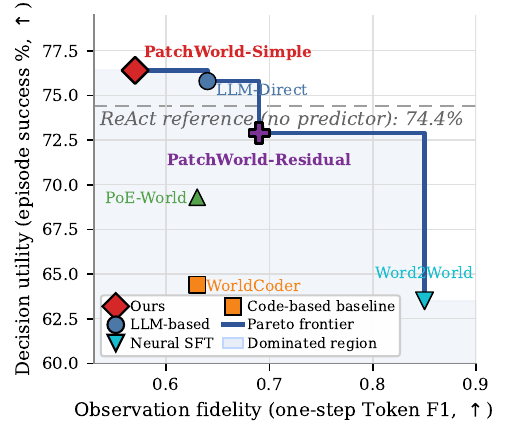}
\vspace{-0.8cm}
\caption{Fidelity-utility Pareto frontier: one-step Token~F1 from
Table~\ref{tab:rq1-onestep} versus live lookahead success from
Table~\ref{tab:appendix-planning-full}.}
\label{fig:pareto}

\Description{Scatter plot of observation fidelity versus decision-making success for compared methods.}
\end{figure}

We embed each world model in the same one-step lookahead loop: a ReAct policy
proposes a default action plus up to four additional LLM candidates; these are
merged with environment-exposed actions, capped at eight total, scored by
predicted next observations, and reranked with a shared LLM selector; the
protocol is detailed in Appendix~\ref{app:rq3-planner}.
Candidate proposal and selection still use LLMs; the comparison isolates the
lookahead predictor among an LLM call, a neural model, and an executable
program.
We evaluate up to 200 held-out episodes per environment with a 30-step cap;
results appear in Tables~\ref{tab:appendix-planning-full}
and~\ref{tab:planning-efficiency} and in Figure~\ref{fig:pareto}.

\begin{table}[t]
\caption{Episode success rate (\%) under the shared one-step lookahead
decision procedure. Average is the unweighted mean over seven environments.
Bold marks the unique best score per column; all-tie columns (Wordle) are left
unbolded.}
\vspace{-0.3cm}
\label{tab:appendix-planning-full}
\centering
\setlength{\tabcolsep}{2.5pt}
\resizebox{\columnwidth}{!}{%
\begin{tabular}{@{}lccccccc|c@{}}
\toprule
\textbf{Method} & Alf. & Baby. & Maze & Sci. & Text. & Web. & Wor. & \textbf{Avg.} \\
\midrule
ReAct              & \textbf{17.5} & 86.7          & 83.3          & 83.5          & 93.8          & 56.0          & 100.0 & 74.4 \\
LLM-Direct         & 14.5          & 88.9          & \textbf{100.0}& 85.0          & 95.8          & 46.0          & 100.0 & 75.7 \\
Word2World         &  9.6          & 75.6          & 83.3          & 66.7          & 86.5          & 22.6          & 100.0 & 63.5 \\
\midrule
POMDP Coder        & 10.0          & \textbf{90.0} & 66.7          & \textbf{100.0}& 90.0          & 65.0          & 100.0 & 74.5 \\
GGP-CWM            & 10.0          & \textbf{90.0} & 50.0          & 80.0          & 85.0          & \textbf{70.0} & 100.0 & 69.3 \\
WorldCoder         &  3.5          & 81.7          & 66.7          & 88.5          & 67.7          & 43.0          & 100.0 & 64.4 \\
PoE-World          &  3.5          & 84.4          & 83.3          & 93.5          & 70.8          & 49.5          & 100.0 & 69.3 \\
\methodsimple{}    &  6.0          & 87.8          & \textbf{100.0}& 86.5          & 95.8          & 58.5          & 100.0 & \textbf{76.4} \\
\methodresidual{}  &  5.5          & 86.1          & 83.3          & 82.5          & \textbf{96.9} & 56.0          & 100.0 & 72.9 \\
\bottomrule
\end{tabular}}
\end{table}

\begin{table}[t]
\caption{Decision-time prediction efficiency. Both rows share the candidate
generator and reranker; only the lookahead predictor
differs. Tokens count lookahead prediction.}
\vspace{-0.3cm}
\label{tab:planning-efficiency}
\centering
\setlength{\tabcolsep}{4pt}
\begin{tabular}{@{}lcc@{}}
\toprule
\textbf{Method} & \textbf{Pred.\ tokens/task} & \textbf{Macro success} \\
\midrule
LLM-Direct & 63{,}897 & 75.7 \\
\methodsimple{} & \textbf{0} & \textbf{76.4} \\
\bottomrule
\end{tabular}
\end{table}

\paragraph{Analysis.}
\textbf{\methodsimple{} attains the best code-based live success at 76.4\%,
exceeding PoE-World at 69.3\% and WorldCoder at 64.4\%, and matching or
exceeding LLM-Direct at 75.7\% while using 0 rather than roughly 64k
lookahead-prediction tokens per task, as reported in
Table~\ref{tab:planning-efficiency}.}
Live decision making evaluates whether belief-conditioned predictions help
choose actions, not only reconstruct text.
Relative to other induced programs,
\methodsimple{} improves macro success by 7.1 points over PoE-World and 12.0
over WorldCoder under the same selector and candidate pool.
Gains concentrate on WebShop and TextCraft, where symbolic updates help reject
infeasible actions; Table~\ref{tab:appendix-planning-full} reports the
per-environment breakdown.
Word2World provides a useful contrast: it has the highest one-step fidelity
at 0.85 F1, yet only 63.5\% macro success, so surface match alone does not
yield better decisions.
An executable belief predictor can therefore replace LLM lookahead without
paying the token cost, and often improves the action choices.

\paragraph{Fidelity is not utility.}
\textbf{Under Qwen3-Coder-480B, \methodresidual{} raises Token~F1 from 0.57
to 0.69 but lowers live success from 76.4\% to 72.9\%.}
Lookahead needs action-relevant contrast about what becomes reachable, not
only closer wording: a prediction can match surface tokens yet still fail to
separate good actions from bad ones under the shared selector.
Appendix~\ref{app:contrast-diagnostics} shows that residual predictions remain
diverse across candidates; the issue is which differences the selector uses,
not that candidates collapse to one string.
Figure~\ref{fig:pareto} illustrates the split:
\methodresidual{} ranks higher on fidelity, while \methodsimple{} ranks
higher on decision utility.
Induced programs appear in Appendix~\ref{app:induced-world-models}.

\subsection{Ablations}
\label{sec:finding-ablation}
\label{sec:finding-cost}

We ablate one component at a time from the full \methodresidual{} pipeline
under a fixed split, backbone, and budget.

\begin{table}[t]
\caption{Component ablation. Avg.\ F1 is test Token~F1; Dec.\ is macro
episode success (\%).}
\vspace{-0.3cm}
\label{tab:rq4-ablation}
\centering
\setlength{\tabcolsep}{3pt}
\begin{tabular}{@{}lcccc@{}}
\toprule
\textbf{Variant} & \textbf{Avg.\ F1} & $\Delta$F1
  & \textbf{Dec.} & $\Delta$Dec \\
\midrule
\textbf{Full} & 0.69 & +0.00
  & 72.9 & +0.0 \\
$-$Residual memory (simple)
  & 0.57 & $-$0.12
  & \textbf{76.4} & +3.5 \\
$-$Repair loop
  & 0.60 & $-$0.09
  & 65.0 & $-$7.9 \\
$-$Contrastive mining
  & 0.49 & $-$0.20
  & 60.0 & $-$12.9 \\
$-$State parsing / text-to-text
  & 0.56 & $-$0.13
  & 48.4 & $-$24.5 \\
\bottomrule
\end{tabular}
\vspace{-0.3cm}
\end{table}

\paragraph{Analysis.}
\textbf{Contrastive mining and belief structure contribute the largest
gains: removing mining reduces F1 by 0.20 and success by 12.9 points, while
removing parsed-state structure reduces success by 24.5 points.}
Without contrastive evidence, the prompt over-represents successful
transitions and under-samples failures, so induction cannot learn the required
conditionals.
Removing the repair loop likewise degrades both metrics, reducing F1 by 0.09
and decision success by 7.9 points, indicating that counterexample-guided
search improves both fidelity and downstream action selection.
As shown in Table~\ref{tab:rq4-ablation}, residual memory again exhibits the
tradeoff between fidelity and utility: removing it lowers F1 by 0.12 but
raises success by 3.5 points.
Taken together, the ablations assign distinct roles to each component: parsed
belief and contrastive mining primarily support decision quality, repair
improves both fidelity and success, and residual memory mainly increases
surface-level match.
Applying the same train-only residual to PoE-World and WorldCoder raises their
F1 from 0.51 to 0.66 and from 0.57 to 0.65, respectively, yet live success
remains 68.3\% and 65.5\%, still below \methodsimple{} at 76.4\%; details
appear in Appendix~\ref{app:pomdp-code-baselines}.
The principal advantage therefore stems from the belief interface and
contrastive repair, rather than from the residual cache alone.

\paragraph{Induction cost.}
\textbf{Induction is incurred offline, averaging 17 to 28 LLM calls and
300k to 503k tokens per environment, after which lookahead uses zero
world-model LLM calls, as reported in Table~\ref{tab:induction-cost-macro}.}
This offline cost yields the decision-time saving in
Table~\ref{tab:planning-efficiency}: once the program is induced, prediction
is ordinary Python execution.
Stronger backbones also reduce validation replay error more sharply:
Qwen by 78.6\%, Mimo by 71.1\%, and DeepSeek by 11.8\%, consistent with
repair being gated by initial synthesis quality.
Details appear in
Appendix~\ref{app:rq5-cost}.

\begin{table}[t]
\caption{Macro induction cost and validation replay-error reduction for
\methodname{} (averages over seven environments).}
\vspace{-0.3cm}
\label{tab:induction-cost-macro}
\centering
\setlength{\tabcolsep}{3pt}
\begin{tabular}{@{}lcccc@{}}
\toprule
\textbf{Backbone} & \textbf{Calls} & \textbf{Rounds}
  & \textbf{Tok./env} & \textbf{Err.\ red.} \\
\midrule
Qwen3-Coder-480B & 28.4 & 6.9 & 452k & 78.6\% \\
Mimo-v2.5 & 25.0 & 6.0 & 503k & 71.1\% \\
DeepSeek-V4-Flash & 17.0 & 4.0 & 300k & 11.8\% \\
\bottomrule
\end{tabular}
\vspace{-0.3cm}
\end{table}

\paragraph{When each variant helps.}
\textbf{\methodresidual{} helps most where train-key coverage is high and
observations are templated, as on AlfWorld, TextCraft, and Wordle.}
Where the symbolic program already covers surface variation, as on Maze and
BabyAI, or where coverage is low and the browser fallback dominates, as on
WebShop, the residual gap shrinks.
After repair, readout and rendering still account for 53\% of remaining
errors on average, as detailed in Appendix~\ref{app:error-analysis}.
These patterns suggest using residual memory for surface-form failures and
the symbolic program when decisions depend on action contrast.

\section{Conclusion}
\label{sec:conclusion}
World models for interactive agents need not live only in neural weights.
\methodname{} turns offline text trajectories into executable belief-state
programs that an LLM proposes, a validator stresses, and local patches improve.
Surface fidelity and decision utility remain related but distinct goals, and
paying for induction once yields a world model that runs without further
model calls at lookahead time.
We therefore argue for treating LLMs as symbolic optimizers of executable
world theories, evaluated on the frontier between fidelity and utility.

\bibliographystyle{ACM-Reference-Format}
\bibliography{references}

\appendix

\section{PatchWorld Repair Loop Details}
\label{app:induce-pseudocode}

Algorithm~\ref{alg:induce} defines the counterexample-guided search.
In implementation, $\sigma(e)$ ranks validator error types: execution and
interface failures first, then missing latent support, then parsed-state or
readout mismatches, then uncategorized failures.
Load failures carry a catastrophic penalty so non-executable programs cannot
beat executable ones: introducing a load error rejects a candidate, and fixing
one is accepted only if it does not create an excessive number of replay
failures.

\section{Dataset Statistics}
\label{app:dataset-stats}

Table~\ref{tab:data-stats} reports per-environment instance, trajectory, and
transition counts under the shared AgentGym collection used throughout.

\begin{table}[h]
\caption{Dataset statistics per environment (instance-level 60/20/20 split).}
\vspace{-0.3cm}
\label{tab:data-stats}
\small
\centering
\begin{tabular*}{\columnwidth}{@{\extracolsep{\fill}}lrrr@{}}
\toprule
\textbf{Env} & \textbf{Instances} & \textbf{Trajectories} & \textbf{Transitions} \\
\midrule
AlfWorld  & 2,620 & 7,860  & 212,695 \\
BabyAI    & 900   & 2,700  & 17,089 \\
Maze      & 26    & 78     & 714 \\
SciWorld  & 2,259 & 6,777  & 146,954 \\
TextCraft & 474   & 1,422  & 10,697 \\
WebShop   & 4,130 & 12,390 & 106,177 \\
Wordle    & 959   & 2,877  & 9,302 \\
\bottomrule
\end{tabular*}
\end{table}

\section{Seed Variance for One-Step Prediction}
\label{app:seed-variance}

We repeat the Qwen3-Coder-480B \methodresidual{} pipeline five times with
independent seeds, refitting both the program and the residual index from
scratch.

\begin{table}[h]
\caption{Five-seed variance for \methodresidual{} (Qwen3-Coder-480B) on the
held-out test set. Mean $\pm$ std over seeds.}
\vspace{-0.3cm}
\label{tab:appendix-seed-variance}
\footnotesize
\centering
\setlength{\tabcolsep}{6pt}
\begin{tabular}{@{}lcc@{}}
\toprule
\textbf{Environment} & \textbf{Token F1} & \textbf{BLEU-4} \\
\midrule
AlfWorld   & $0.6982 \pm 0.0205$ & $0.4665 \pm 0.0188$ \\
BabyAI     & $0.6878 \pm 0.0247$ & $0.4691 \pm 0.0264$ \\
Maze       & $0.9663 \pm 0.0089$ & $0.9091 \pm 0.0142$ \\
SciWorld   & $0.5618 \pm 0.0291$ & $0.4761 \pm 0.0253$ \\
TextCraft  & $0.7104 \pm 0.0186$ & $0.4805 \pm 0.0172$ \\
WebShop    & $0.4985 \pm 0.0224$ & $0.2603 \pm 0.0179$ \\
Wordle     & $0.7363 \pm 0.0102$ & $0.4324 \pm 0.0176$ \\
\midrule
\textbf{Macro} & $\mathbf{0.6942 \pm 0.0192}$ & $\mathbf{0.4991 \pm 0.0231}$ \\
\bottomrule
\end{tabular}
\end{table}

\textbf{Macro seed std ($\approx 0.02$ Token~F1) is small relative to the gap
between \methodresidual{} and the strongest non-residual baseline
($\approx 0.06$).}

\section{BERTScore Consistency with Token F1 and BLEU-4}
\label{app:bertscore-consistency}

We score held-out predictions from the five program-based one-step
configurations with BERTScore~\citep{zhang2020bertscore}
(\texttt{roberta-large}, English baseline, no re-scaling), replaying
persisted programs with no new LLM calls. Token~F1 and BLEU-4 in
Table~\ref{tab:appendix-bertscore-macro} repeat the main-table macros for
side-by-side comparison; BERTScore-F1 is the additional semantic metric.

\begin{table}[h]
\caption{Macro BERTScore-F1 on replayed held-out predictions, with main-table
Token~F1 / BLEU-4 from Table~\ref{tab:rq1-onestep} for reference.
Bold marks the best score.}
\vspace{-0.3cm}
\label{tab:appendix-bertscore-macro}
\footnotesize
\centering
\setlength{\tabcolsep}{3pt}
\resizebox{\columnwidth}{!}{%
\begin{tabular}{@{}lccc@{}}
\toprule
\textbf{Configuration} & \textbf{TokF1} & \textbf{BLEU4} & \textbf{BERT-F1} \\
\midrule
\methodsimple{} + Mimo-v2.5            & 0.60 & 0.38 & 0.916 \\
\methodsimple{} + Qwen3-Coder-480B     & 0.57 & 0.34 & 0.934 \\
\methodresidual{} + DeepSeek-V4-Flash  & \textbf{0.70} & \textbf{0.50} & 0.948 \\
\methodresidual{} + Qwen3-Coder-480B   & 0.69 & \textbf{0.50} & 0.948 \\
\methodresidual{} + Mimo-v2.5          & \textbf{0.70} & 0.49 & \textbf{0.950} \\
\bottomrule
\end{tabular}}
\end{table}

\textbf{BERTScore agrees with the main fidelity conclusion: residual variants
beat simple ones, and within residual the three backbones remain nearly tied
(0.948--0.950).}
BERTScore's compressed range can mask exact failures (a zero Token~F1 cell
can still score near $0.85$), so we keep Token~F1 and BLEU-4 as primary
fidelity metrics.

\section{Residual Memory Coverage Diagnostic}
\label{app:residual-coverage}

Table~\ref{tab:residual-coverage} measures how often an exact normalized
train (observation, action) key appears in held-out test transitions.
A retrieval-only diagnostic returns the memorized majority next observation
on a hit and scores misses as zero Token~F1.

\textbf{The residual is high precision but low coverage (macro hit rate 0.33),
so most held-out transitions still rely on the executable program.}

\begin{table}[h]
\caption{Exact train-key residual coverage on the held-out test split.
Hit F1 is Token F1 on cache hits; All F1 counts misses as zero.}
\vspace{-0.3cm}
\label{tab:residual-coverage}
\footnotesize
\centering
\setlength{\tabcolsep}{3.5pt}
\begin{tabular}{@{}lccc@{}}
\toprule
\textbf{Env.} & \textbf{Hit rate} & \textbf{Hit F1} & \textbf{All F1} \\
\midrule
AlfWorld   & 0.20 & 0.94 & 0.18 \\
BabyAI     & 0.12 & 0.95 & 0.12 \\
Maze       & 0.87 & 1.00 & 0.87 \\
SciWorld   & 0.45 & 0.99 & 0.44 \\
TextCraft  & 0.42 & 0.99 & 0.41 \\
WebShop    & 0.00 & 0.00 & 0.00 \\
Wordle     & 0.26 & 0.95 & 0.24 \\
\midrule
\textbf{Macro} & \textbf{0.33} & \textbf{0.83} & \textbf{0.32} \\
\bottomrule
\end{tabular}
\end{table}

\section{Per-Environment Rollout Results}
\label{app:rq2-full-rollout}

Table~\ref{tab:appendix-rollout-full} expands the macro rollout summary in
Table~\ref{tab:rq2-macro-rollout}.
Cells use episode-level filtering (episodes must be long enough for the
horizon), so $t{=}1$ is not directly comparable to Table~\ref{tab:rq1-onestep}.

\begin{table}[h]
\caption{Rollout Token F1 by environment and horizon. Bold marks the best
code-based score per cell; \methodname{} variants share the
Qwen3-Coder-480B backbone.}
\vspace{-0.3cm}
\label{tab:appendix-rollout-full}
\scriptsize
\centering
\setlength{\tabcolsep}{3pt}
\begin{tabular}{@{}llcccc@{}}
\toprule
\textbf{Env.} & \textbf{Method} & \textbf{$t{=}1$} & \textbf{$t{=}2$} & \textbf{$t{=}3$} & \textbf{$t{=}5$} \\
\midrule
\multirow{4}{*}{AlfWorld}
  & WorldCoder         & 0.62 & 0.59 & 0.57 & 0.56 \\
  & PoE-World          & 0.62 & 0.55 & 0.50 & 0.46 \\
  & \methodsimple{}    & 0.36 & 0.31 & 0.28 & 0.27 \\
  & \methodresidual{}  & \textbf{0.69} & \textbf{0.62} & \textbf{0.58} & \textbf{0.57} \\
\midrule
\multirow{4}{*}{BabyAI}
  & WorldCoder         & 0.78 & 0.75 & \textbf{0.74} & \textbf{0.74} \\
  & PoE-World          & 0.77 & 0.71 & 0.68 & 0.66 \\
  & \methodsimple{}    & \textbf{0.83} & \textbf{0.77} & 0.69 & 0.63 \\
  & \methodresidual{}  & 0.69 & 0.65 & 0.62 & 0.62 \\
\midrule
\multirow{4}{*}{Maze}
  & WorldCoder         & 0.83 & 0.81 & 0.79 & 0.78 \\
  & PoE-World          & 0.83 & 0.78 & 0.74 & 0.70 \\
  & \methodsimple{}    & 0.80 & 0.74 & 0.66 & 0.61 \\
  & \methodresidual{}  & \textbf{0.97} & \textbf{0.94} & \textbf{0.90} & \textbf{0.85} \\
\midrule
\multirow{4}{*}{SciWorld}
  & WorldCoder         & 0.40 & 0.37 & 0.35 & 0.34 \\
  & PoE-World          & 0.39 & 0.34 & 0.31 & 0.30 \\
  & \methodsimple{}    & \textbf{0.57} & \textbf{0.49} & 0.41 & 0.39 \\
  & \methodresidual{}  & 0.55 & \textbf{0.49} & \textbf{0.44} & \textbf{0.41} \\
\midrule
\multirow{4}{*}{TextCraft}
  & WorldCoder         & \textbf{0.88} & \textbf{0.85} & \textbf{0.83} & \textbf{0.82} \\
  & PoE-World          & 0.78 & 0.73 & 0.70 & 0.69 \\
  & \methodsimple{}    & 0.28 & 0.22 & 0.18 & 0.15 \\
  & \methodresidual{}  & 0.71 & 0.66 & 0.64 & 0.63 \\
\midrule
\multirow{4}{*}{WebShop}
  & WorldCoder         & \textbf{0.58} & \textbf{0.55} & \textbf{0.52} & \textbf{0.50} \\
  & PoE-World          & \textbf{0.58} & 0.52 & 0.46 & 0.43 \\
  & \methodsimple{}    & 0.53 & 0.46 & 0.38 & 0.36 \\
  & \methodresidual{}  & 0.50 & 0.48 & 0.46 & 0.45 \\
\midrule
\multirow{4}{*}{Wordle}
  & WorldCoder         & 0.31 & 0.29 & 0.28 & 0.28 \\
  & PoE-World          & 0.43 & 0.38 & 0.32 & 0.30 \\
  & \methodsimple{}    & 0.58 & 0.49 & 0.41 & 0.39 \\
  & \methodresidual{}  & \textbf{0.72} & \textbf{0.65} & \textbf{0.55} & \textbf{0.50} \\
\bottomrule
\end{tabular}
\end{table}

\paragraph{Pattern.}
\textbf{\methodresidual{} decays more slowly where templated readouts dominate,
while \methodsimple{} can be stronger at short horizons yet drift faster on
token metrics.}
This matches the main-text split between surface fidelity and decision utility:
symbolic state can remain useful even when rendered text leaves the simulator
template.

\section{Decision Protocol}
\label{app:rq3-planner}

Every world model uses the same one-step lookahead procedure; only the
predictor module differs:

\begin{enumerate}
  \item \textbf{Belief update.} On observation $o_t$, the world model
        ingests $(o_{t-1},a_{t-1},o_t)$ and updates its internal state. For
        \methodname{}, this calls \code{correct_belief}; for
        WorldCoder/PoE-World, the program's transition; for LLM-Direct, an
        ICL prompt that conditions on the last $k{=}3$ transitions.
  \item \textbf{Candidate generation.} A ReAct policy proposes a default
        action $a^{\text{default}}$ plus up to four additional LLM
        candidates. These are merged with the environment's exposed action
        API, then deduplicated and capped at eight total candidates
        (including the default).
  \item \textbf{Lookahead rollout.} For each candidate $a^{(i)}$, the world
        model predicts $\hat{o}_{t+1}^{(i)}$. No multi-step rollout is used
        in this study; the lookahead depth is exactly one.
  \item \textbf{Gated reranking.} A shared Qwen3-Coder-480B selector scores
        $(o_t, a^{(i)}, \hat{o}_{t+1}^{(i)})$ tuples. The decision procedure overrides
        $a^{\text{default}}$ only when the world-model predictions provide a
        usable contrast and the selected action has clear positive evidence
        relative to the default (or avoids an explicitly bad predicted
        outcome). Otherwise it falls back to $a^{\text{default}}$. This avoids
        penalizing strong reactive baselines when the world model adds no
        reliable signal.
  \item \textbf{Termination.} Episodes terminate on environment success,
        environment failure, or a 30-step cap, whichever comes first. Each
        environment uses up to 200 held-out instances.
\end{enumerate}

The selector, candidate cap, and step cap are identical across all rows in
Figure~\ref{fig:pareto} and Table~\ref{tab:appendix-planning-full}.
Table~\ref{tab:planning-efficiency} reports lookahead-prediction token cost
for LLM-Direct versus \methodsimple{}.
Deeper search, reward models, and distributional rollouts are orthogonal
extensions.

\section{Decision-Making Uncertainty}
\label{app:rq3-planning-uncertainty}

Table~\ref{tab:appendix-planning-uncertainty} reports macro live success as a
five-run mean $\pm$ standard deviation under the same unweighted estimator as
Table~\ref{tab:appendix-planning-full}.

\begin{table}[h]
\caption{Macro episode success as five-run mean $\pm$ standard deviation.}
\vspace{-0.3cm}
\label{tab:appendix-planning-uncertainty}
\footnotesize
\centering
\setlength{\tabcolsep}{5pt}
\begin{tabular}{@{}lc@{}}
\toprule
\textbf{Method} & \textbf{Macro success (\%)} \\
\midrule
ReAct              & $74.4 \pm 2.3$ \\
LLM-Direct         & $75.7 \pm 0.8$ \\
Word2World         & $63.5 \pm 1.2$ \\
WorldCoder         & $64.4 \pm 2.9$ \\
PoE-World          & $69.3 \pm 2.4$ \\
\methodsimple{}    & $\mathbf{76.4 \pm 0.8}$ \\
\methodresidual{}  & $72.9 \pm 2.3$ \\
\bottomrule
\end{tabular}
\end{table}

\textbf{Code-baseline gaps exceed these standard deviations, while
\methodsimple{} remains within error of---or ahead of---LLM-Direct.}

\section{Cross-Action Contrast Diagnostics}
\label{app:contrast-diagnostics}

Table~\ref{tab:contrast-diagnostics} checks whether residual memory collapses
candidate predictions into identical strings during live lookahead.
We normalize whitespace and case, discard empty predictions, and report the
share of steps with fewer than two distinct usable predictions, the mean
number of distinct predictions, and mean pairwise SequenceMatcher distance.

\begin{table}[h]
\caption{Cross-action contrast from saved live decision-making logs.
Identical is the share of steps with fewer than two distinct usable
predictions.}
\vspace{-0.3cm}
\label{tab:contrast-diagnostics}
\footnotesize
\centering
\setlength{\tabcolsep}{4pt}
\begin{tabular}{@{}lccc@{}}
\toprule
\textbf{Model} & \textbf{Identical (\%)} & \textbf{Unique preds.} & \textbf{Pairwise dist.} \\
\midrule
\methodsimple{}   & 6.4 & 4.73 & 0.42 \\
\methodresidual{} & 5.7 & 4.87 & 0.44 \\
\bottomrule
\end{tabular}
\end{table}

\textbf{Residual predictions are at least as diverse as \methodsimple{}, so
the live-success drop is not from identical retrieved strings across
candidates.}

\section{Component Ablation Details}
\label{app:component-ablation}

Table~\ref{tab:rq4-ablation} variants, relative to full \methodresidual{}:
$-$Residual memory reduces to \methodsimple{}; $-$Repair loop sets $R{=}0$;
$-$Contrastive mining replaces mining with uniform sampling; and
$-$State parsing / text-to-text removes parsed-state structure and forces a
direct text-to-text predictor.

\section{Executable Code-World-Model Baselines under Text-Only Supervision}
\label{app:pomdp-code-baselines}

We adapt POMDP Coder~\citep{curtis2025pomdp} and
GGP-CWM~\citep{lehrach2025code} to AgentGym's text-only supervision
(observations, actions, rewards; no simulator hidden state).
POMDP Coder keeps its four-model interface with coverage-based repair over
text-state surrogates; GGP-CWM uses a paper-faithful helper decomposition
(transition, legal action, terminal, reward, value, inference).

\begin{table}[h]
\caption{Executable code world models under text-only AgentGym supervision.
F1 is test Token~F1; Dec.\ is live success (\%).}
\vspace{-0.3cm}
\label{tab:pomdp-code-baselines}
\footnotesize
\centering
\setlength{\tabcolsep}{3pt}
\begin{tabular}{@{}lcc@{}}
\toprule
\textbf{Method} & \textbf{F1} & \textbf{Dec.} \\
\midrule
\methodsimple{} & 0.57 & \textbf{76.4} \\
\methodresidual{} & \textbf{0.69} & 72.9 \\
POMDP Coder & 0.50 & 74.5 \\
GGP-CWM & 0.29 & 69.3 \\
WorldCoder & 0.63 & 64.4 \\
PoE-World & 0.63 & 69.3 \\
\bottomrule
\end{tabular}
\end{table}

\textbf{\methodresidual{} attains the highest offline fidelity and
\methodsimple{} the highest live success among these executable methods:
belief abstraction plus repair fits text-only logs better than substituting
parsed text for a hidden state.}

\begin{table}[h]
\caption{Per-environment test Token~F1 for POMDP/code baselines.}
\vspace{-0.3cm}
\label{tab:pomdp-code-baselines-env}
\footnotesize
\centering
\setlength{\tabcolsep}{2pt}
\resizebox{\columnwidth}{!}{%
\begin{tabular}{@{}lccc@{}}
\toprule
\textbf{Env} & \methodresidual{} & POMDP Coder & GGP-CWM \\
\midrule
AlfWorld & 0.70 & 0.67 & 0.26 \\
BabyAI & 0.69 & 0.59 & 0.42 \\
Maze & 0.97 & 0.85 & 0.71 \\
SciWorld & 0.56 & 0.40 & 0.00 \\
TextCraft & 0.71 & 0.73 & 0.24 \\
WebShop & 0.50 & 0.24 & 0.36 \\
Wordle & 0.72 & 0.00 & 0.00 \\
\midrule
Avg. & \textbf{0.69} & 0.50 & 0.29 \\
\bottomrule
\end{tabular}}
\end{table}

Table~\ref{tab:residual-on-baselines} wraps the same train-only residual onto
transition-code baselines.

\begin{table}[h]
\caption{Residual-memory wrapper on programmatic baselines.
PatchWorld uses the Qwen3-Coder-480B Simple$\to$Residual pair from
Tables~\ref{tab:rq1-onestep} and~\ref{tab:appendix-planning-full}.
PoE-World / WorldCoder wrap the DeepSeek-V4-Flash programs from
Table~\ref{tab:rq1-onestep} (Base F1 $0.51$ / $0.57$) with the same
train-only residual; Base Dec.\ matches the shared planning table, and
$+$Res.\ Dec.\ is from the residualized live runs.}
\vspace{-0.3cm}
\label{tab:residual-on-baselines}
\footnotesize
\centering
\setlength{\tabcolsep}{2.5pt}
\begin{tabular}{@{}lcccc@{}}
\toprule
\textbf{Method} & \textbf{Base F1} & \textbf{+Res.\ F1}
  & \textbf{Base Dec.} & \textbf{+Res.\ Dec.} \\
\midrule
\methodsimple{}$\to$\methodresidual{} & 0.57 & 0.69 & 76.4 & 72.9 \\
PoE-World (DeepSeek) & 0.51 & 0.66 & 69.3 & 68.3 \\
WorldCoder (DeepSeek) & 0.57 & 0.65 & 64.4 & 65.5 \\
\bottomrule
\end{tabular}
\end{table}

\textbf{Residual memory raises F1 across methods---including non-PatchWorld
programs---but does not close the decision gap: even residualized PoE-World
(68.3\%) and WorldCoder (65.5\%) remain below \methodsimple{} (76.4\%), so
the core advantage remains belief structure plus repair, not the cache alone.}
Table~\ref{tab:wordle-position} is a Wordle-only secondary check; the unified
Token~F1 / BLEU protocol remains primary.

\begin{table}[h]
\caption{Wordle position-sensitive feedback on held-out validation transitions.}
\vspace{-0.3cm}
\label{tab:wordle-position}
\footnotesize
\centering
\setlength{\tabcolsep}{3pt}
\begin{tabular}{@{}lccc@{}}
\toprule
\textbf{Method} & \textbf{Seq.\ exact} & \textbf{Slot acc.} & \textbf{Pos.\ F1} \\
\midrule
\methodresidual{} & 26.7 & 62.9 & 54.4 \\
\methodsimple{} & 7.5 & 50.0 & 39.0 \\
PoE-World & 5.9 & 42.0 & 34.2 \\
WorldCoder & 5.9 & 42.0 & 34.2 \\
\bottomrule
\end{tabular}
\end{table}

\section{Detailed Induction Cost and Replay-Error Reduction}
\label{app:rq5-cost}

Macro averages appear in Table~\ref{tab:induction-cost-macro}.
Tables~\ref{tab:rq5-efficiency} and~\ref{tab:rq5-patchworld-optimization}
give per-environment call/token/round counts and validation replay-error
reduction under the \loopname{}.

\begin{table}[h]
\caption{Per-environment induction cost for \methodname{}.
Rounds are attempted repair iterations before convergence, budget ($R{=}15$),
or hill-climbing rejection. Tokens rounded to thousands.}
\vspace{-0.3cm}
\label{tab:rq5-efficiency}
\scriptsize
\centering
\setlength{\tabcolsep}{2pt}
\resizebox{\columnwidth}{!}{%
\begin{tabular}{@{}ll rrrrl@{}}
\toprule
\textbf{Backbone} & \textbf{Env}
  & \textbf{Calls} & \textbf{In tok.} & \textbf{Out tok.}
  & \textbf{Rounds} & \textbf{Conv.?} \\
\midrule
Qwen3-Coder-480B & AlfWorld  & 25 & 245.1k & 97.1k  & 6  & No \\
Qwen3-Coder-480B & BabyAI    & 29 & 446.5k & 97.0k  & 7  & No \\
Qwen3-Coder-480B & Maze      & 17 & 128.3k & 37.6k  & 4  & No \\
Qwen3-Coder-480B & SciWorld  & 29 & 219.6k & 126.2k & 7  & No \\
Qwen3-Coder-480B & TextCraft & 21 & 153.6k & 84.3k  & 5  & No \\
Qwen3-Coder-480B & WebShop   & 45 & 1002.0k & 181.2k & 11 & No \\
Qwen3-Coder-480B & Wordle    & 33 & 227.5k & 115.3k & 8  & No \\
\cmidrule(l){2-7}
Qwen3-Coder-480B & Avg.      & 28.4 & 346.1k & 105.5k & 6.9 & No \\
\midrule
Mimo-v2.5 & AlfWorld  & 25 & 230.8k & 256.2k & 6  & No \\
Mimo-v2.5 & BabyAI    & 17 & 331.0k & 173.8k & 4  & No \\
Mimo-v2.5 & Maze      & 17 & 78.5k  & 99.2k  & 4  & No \\
Mimo-v2.5 & SciWorld  & 29 & 377.7k & 366.7k & 7  & No \\
Mimo-v2.5 & TextCraft & 45 & 303.6k & 390.2k & 11 & No \\
Mimo-v2.5 & WebShop   & 13 & 297.6k & 130.2k & 3  & No \\
Mimo-v2.5 & Wordle    & 29 & 209.0k & 277.3k & 7  & No \\
\cmidrule(l){2-7}
Mimo-v2.5 & Avg.      & 25.0 & 261.2k & 241.9k & 6.0 & No \\
\midrule
DeepSeek-V4-Flash & AlfWorld  & 25 & 208.8k & 98.1k & 6 & No \\
DeepSeek-V4-Flash & BabyAI    & 13 & 304.7k & 58.1k & 3 & No \\
DeepSeek-V4-Flash & Maze      & 17 & 134.5k & 39.5k & 4 & No \\
DeepSeek-V4-Flash & SciWorld  & 17 & 207.3k & 84.8k & 4 & No \\
DeepSeek-V4-Flash & TextCraft & 13 & 94.1k  & 37.0k & 3 & No \\
DeepSeek-V4-Flash & WebShop   & 21 & 571.5k & 79.5k & 5 & No \\
DeepSeek-V4-Flash & Wordle    & 13 & 112.0k & 69.3k & 3 & No \\
\cmidrule(l){2-7}
DeepSeek-V4-Flash & Avg. & 17.0 & 233.3k & 66.6k & 4.0 & No \\
\bottomrule
\end{tabular}}
\end{table}

\begin{table}[h]
\caption{\loopname{} replay-error reduction on validation trajectories
(relative to the initial generated program).}
\vspace{-0.3cm}
\label{tab:rq5-patchworld-optimization}
\scriptsize
\centering
\setlength{\tabcolsep}{2pt}
\resizebox{\columnwidth}{!}{%
\begin{tabular}{@{}ll rrrr@{}}
\toprule
\textbf{Backbone} & \textbf{Env}
  & \textbf{Init. err.} & \textbf{Final err.}
  & \textbf{$\Delta$ err.} & \textbf{Red.} \\
\midrule
Qwen3-Coder-480B & AlfWorld  & 7866  & 1059 & 6807  & 86.5\% \\
Qwen3-Coder-480B & BabyAI    & 2102  & 951  & 1151  & 54.8\% \\
Qwen3-Coder-480B & Maze      & 507   & 305  & 202   & 39.8\% \\
Qwen3-Coder-480B & SciWorld  & 3472  & 24   & 3448  & 99.3\% \\
Qwen3-Coder-480B & TextCraft & 2028  & 404  & 1624  & 80.1\% \\
Qwen3-Coder-480B & WebShop   & 1060  & 552  & 508   & 47.9\% \\
Qwen3-Coder-480B & Wordle    & 475   & 446  & 29    & 6.1\% \\
\cmidrule(l){2-6}
Qwen3-Coder-480B & Total     & 17510 & 3741 & 13769 & 78.6\% \\
\midrule
Mimo-v2.5 & AlfWorld  & 5720  & 1430 & 4290  & 75.0\% \\
Mimo-v2.5 & BabyAI    & 1564  & 518  & 1046  & 66.9\% \\
Mimo-v2.5 & Maze      & 508   & 20   & 488   & 96.1\% \\
Mimo-v2.5 & SciWorld  & 6113  & 1788 & 4325  & 70.8\% \\
Mimo-v2.5 & TextCraft & 334   & 162  & 172   & 51.5\% \\
Mimo-v2.5 & WebShop   & 460   & 460  & 0     & 0.0\% \\
Mimo-v2.5 & Wordle    & 462   & 10   & 452   & 97.8\% \\
\cmidrule(l){2-6}
Mimo-v2.5 & Total     & 15161 & 4388 & 10773 & 71.1\% \\
\midrule
DeepSeek-V4-Flash & AlfWorld  & 723  & 675  & 48  & 6.6\% \\
DeepSeek-V4-Flash & BabyAI    & 479  & 479  & 0   & 0.0\% \\
DeepSeek-V4-Flash & Maze      & 318  & 142  & 176 & 55.3\% \\
DeepSeek-V4-Flash & SciWorld  & 2665 & 2200 & 465 & 17.4\% \\
DeepSeek-V4-Flash & TextCraft & 164  & 164  & 0   & 0.0\% \\
DeepSeek-V4-Flash & WebShop   & 1266 & 1232 & 34  & 2.7\% \\
DeepSeek-V4-Flash & Wordle    & 517  & 517  & 0   & 0.0\% \\
\cmidrule(l){2-6}
DeepSeek-V4-Flash & Total     & 6132 & 5409 & 723 & 11.8\% \\
\bottomrule
\end{tabular}}
\end{table}

\textbf{Replay-error reduction is gated by initial synthesis quality: strong
for Qwen and Mimo, small for DeepSeek, and no environment reaches exact
convergence within $R{=}15$.}
Longer-horizon environments also tend to consume more calls and rounds.

\section{Validator Error Analysis}
\label{app:error-analysis}

We aggregate typed validator tags---parser, transition, readout/render, or
unhandled action---over the \methodresidual{} (Qwen3-Coder-480B) post-repair
validation log.

\begin{table}[h]
\caption{Post-repair validation errors by typed cluster (rows sum to 100\%).
Unhandled covers rejected or raising actions.}
\vspace{-0.3cm}
\label{tab:appendix-error-clusters}
\footnotesize
\centering
\setlength{\tabcolsep}{4pt}
\begin{tabular}{@{}lcccc@{}}
\toprule
\textbf{Env.} & \textbf{Parser} & \textbf{Transition} & \textbf{Readout} & \textbf{Unhandled} \\
\midrule
AlfWorld   &  6\% & 18\% & \textbf{63\%} & 13\% \\
BabyAI     &  4\% & 31\% & 49\%          & 16\% \\
Maze       &  3\% & 22\% & 60\%          & 15\% \\
SciWorld   &  9\% & 20\% & \textbf{58\%} & 13\% \\
TextCraft  &  5\% & \textbf{47\%} & 32\% & 16\% \\
WebShop    &  7\% & \textbf{44\%} & 36\% & 13\% \\
Wordle     &  3\% & 12\% & \textbf{75\%} & 10\% \\
\midrule
\textbf{Macro} & 5\% & 28\% & 53\% & 14\% \\
\bottomrule
\end{tabular}
\end{table}

\paragraph{Implication.}
\textbf{Readout/rendering is 53\% of remaining errors on average and dominates
templated environments, which is where residual memory helps most.}
Transition errors concentrate in TextCraft and WebShop, where compound rules
stress program expressivity.
Natural next steps are fill-in renderer templates and validator checks beyond
exact string match.

\section{Qualitative PatchWorld Repair Example}
\label{app:qualitative-repair}

Figure~\ref{fig:qualitative-repair} shows a Maze patch that fixed over-eager
terminal parsing: the initial program treated any observation containing
\texttt{Success} as done, but the first Maze observation embeds an
instructional example. After \texttt{move right}, the validator showed a
coordinate update with a wall above, while the program predicted terminal
\texttt{Success} or dropped walls. The accepted patch tightened the terminal
check and preserved walls on blocked moves (127$\to$109 validation errors).

\section{Induced World Models by Environment}
\label{app:induced-world-models}

Figures~\ref{fig:wm-alfworld}--\ref{fig:wm-wordle} list the induced Python
programs for each environment under the shared \code{BaseWorldModel}
interface (\code{parse_observation}, \code{init_belief},
\code{correct_belief}, \code{predict_belief}, \code{readout_observation}).
Source files ship in the release under \code{generated_world_models/}.

\section{Limitations}
\label{app:limitations}
\label{sec:limitations}

We isolate the world model with one-step lookahead rather than a learned
controller, value function, or deep search.
Because programs predict observations but not rewards, live decision making
still uses a shared LLM selector; controlling that selector across methods
does not remove interactions with prediction style.
We induce deterministic belief programs, not distributional
transition/reward models, and one model per environment, so we do not
measure cross-task transfer or expected-utility planning under
underdetermined outcomes.
Decision-making gains are environment-dependent (e.g., AlfWorld can trail
the ReAct default), the benchmark is language-only, and interpretability
claims rest on inspectable programs and execution diagnostics rather than
user studies.

\section{Use of AI Assistants}
We used AI-based tools for grammar fixing and polishing the writing of the manuscript.

\section{Risks}
\label{app:risks}

An induced executable model can look interpretable while still encoding wrong
rules, missing hidden state, or gaps inherited from limited trajectories.
Without checks, such errors can yield confident but incorrect decisions.
Generated models should be sandboxed, reviewed, and treated as diagnostic or
assistive tools in higher-stakes settings, not as trusted simulators.

\clearpage
\onecolumn
\renewcommand{\lstlistingname}{Figure}
\setcounter{lstlisting}{4}

\begin{lstlisting}[
  style=worldmodelcode,
  caption={\textbf{AlfWorld induced world model.} Executable program for
  household navigation with receptacle/object tracking, inventory updates, and
  action-conditioned readout.},
  label={fig:wm-alfworld},
]
from abductworld.worldmodel_base import BaseWorldModel
import re
from collections import defaultdict

class AlfworldWorldModel(BaseWorldModel):
    def __init__(self):
        super().__init__()
        self.receptacles = set()
        self.objects = set()
        self.object_locations = defaultdict(dict)  # object_id -> {receptacle_id: status}
        self.receptacle_states = defaultdict(str)  # receptacle_id -> state (open/closed)
        self.inventory = set()
        self.current_location = None
        self.valid_actions = {
            "go to": ["go to <receptacle>"],
            "take": ["take <object> from <receptacle>"],
            "move": ["move <object> to <receptacle>"],
            "examine": ["examine <object>"],
            "look": ["look"],
            "inventory": ["inventory"],
            "open": ["open <receptacle>"],
            "close": ["close <receptacle>"],
            "use": ["use <object>"],
            "heat": ["heat <object> with <receptacle>"],
            "clean": ["clean <object> with <receptacle>"]
        }

    def parse_observation(self, obs_text: str) -> dict:
        """Parse observation text into structured data"""
        parsed = {
            "location": None,
            "objects": [],
            "receptacles": [],
            "inventory": [],
            "message": obs_text.strip()
        }
        
        # Extract location
        location_match = re.search(r"You are in the middle of a room\. Looking quickly around you, you see (.+)$", obs_text)
        if location_match:
            items = location_match.group(1).split(", ")
            for item in items:
                if " " in item:
                    name, num = item.rsplit(" ", 1)
                    parsed["receptacles"].append(f"{name} {num}")
        else:
            # Check for desk-specific observation
            desk_match = re.search(r"You arrive at ([^\.]+)\. On the ([^,]+), you see (.+)", obs_text)
            if desk_match:
                parsed["location"] = desk_match.group(1)
                objects_str = desk_match.group(3)
                if objects_str != "nothing":
                    object_list = objects_str.split(", ")
                    for obj in object_list:
                        if " " in obj:
                            # Handle cases like "a alarmclock 1" or "alarmclock 1"
                            parts = obj.replace("a ", "").strip().split(" ")
                            if len(parts) >= 2:
                                obj_name = " ".join(parts[:-1])
                                obj_id = parts[-1]
                                parsed["objects"].append(f"{obj_name} {obj_id}")
        
        # Check for inventory
        if "You are not carrying anything" in obs_text:
            parsed["inventory"] = []
        elif "You are carrying:" in obs_text:
            inv_match = re.search(r"You are carrying: (.+)", obs_text)
            if inv_match:
                items = inv_match.group(1).split(", ")
                parsed["inventory"] = [item.strip() for item in items]
                
        return parsed

    def init_belief(self):
        """Initialize belief state with latent support"""
        return {
            "location": None,
            "receptacles": set(),
            "objects": set(),
            "object_locations": defaultdict(dict),
            "receptacle_states": defaultdict(str),
            "inventory": set(),
            "last_action": None,
            # Adding latent belief fields to satisfy validator
            "latent_variables": {},
            "facts": set(),
            "hypotheses": [],
            "frontier": set(),
            "hidden_state": {}
        }

    def correct_belief(self, belief_prior, obs_text: str):
        """Update belief state based on observation"""
        belief = belief_prior.copy()
        # Preserve latent fields
        if "latent_variables" not in belief:
            belief["latent_variables"] = {}
        if "facts" not in belief:
            belief["facts"] = set()
        if "hypotheses" not in belief:
            belief["hypotheses"] = []
        if "frontier" not in belief:
            belief["frontier"] = set()
        if "hidden_state" not in belief:
            belief["hidden_state"] = {}
            
        parsed = self.parse_observation(obs_text)
        
        # Update location
        if parsed["location"]:
            belief["location"] = parsed["location"]
            
        # Update inventory
        belief["inventory"] = set(parsed["inventory"])
        
        # Update object locations from "On the X, you see..." patterns
        on_pattern = r"On the ([^,]+), you see (.+)"
        on_matches = re.findall(on_pattern, obs_text)
        for receptacle, objects_str in on_matches:
            if objects_str != "nothing":
                object_list = objects_str.split(", ")
                for obj in object_list:
                    obj_clean = obj.replace("a ", "").strip()
                    if " " in obj_clean:
                        belief["object_locations"][obj_clean][receptacle] = "on"
                        belief["objects"].add(obj_clean)
        
        # Update receptacle states
        if "is open" in obs_text:
            open_matches = re.findall(r"The ([^ ]+ [0-9]+) is open", obs_text)
            for receptacle in open_matches:
                belief["receptacle_states"][receptacle] = "open"
                
        if "is closed" in obs_text:
            closed_matches = re.findall(r"The ([^ ]+ [0-9]+) is closed", obs_text)
            for receptacle in closed_matches:
                belief["receptacle_states"][receptacle] = "closed"
                
        return belief

    def predict_belief(self, belief, action: str):
        """Predict next belief state given action"""
        # Create a deep copy that preserves all fields including latent ones
        next_belief = {}
        for key, value in belief.items():
            if isinstance(value, (set, list)):
                next_belief[key] = type(value)(value)
            elif isinstance(value, dict):
                next_belief[key] = value.copy()
            else:
                next_belief[key] = value
        
        next_belief["last_action"] = action
        
        # Parse action
        action = action.strip().lower()
        
        if action.startswith("go to "):
            location = action[6:]  # Remove "go to "
            next_belief["location"] = location
            
        elif action.startswith("take "):
            # take <object> from <receptacle>
            match = re.match(r"take (.+) from (.+)", action)
            if match:
                obj, receptacle = match.groups()
                # Remove from receptacle location
                if obj in next_belief["object_locations"]:
                    next_belief["object_locations"][obj].pop(receptacle, None)
                # Add to inventory
                next_belief["inventory"].add(obj)
                
        elif action.startswith("move "):
            # move <object> to <receptacle>
            match = re.match(r"move (.+) to (.+)", action)
            if match:
                obj, receptacle = match.groups()
                # Remove from inventory if present
                next_belief["inventory"].discard(obj)
                # Add to receptacle location
                next_belief["object_locations"][obj][receptacle] = "on"
                
        elif action.startswith("open "):
            receptacle = action[5:]  # Remove "open "
            next_belief["receptacle_states"][receptacle] = "open"
            
        elif action.startswith("close "):
            receptacle = action[6:]  # Remove "close "
            next_belief["receptacle_states"][receptacle] = "closed"
            
        return next_belief

    def readout_observation(self, belief, action: str = "") -> str:
        """Generate observation text from belief state"""
        # For simplicity, return a generic success message for now
        # In a full implementation, this would reconstruct the observation
        # based on the belief state and action taken
        if belief["last_action"]:
            action_type = belief["last_action"].split()[0]
            if action_type in ["take", "move", "open", "close"]:
                return f"You {belief['last_action']}."
            elif action_type == "go":
                return f"You arrive at {belief['location']}."
            elif action_type == "examine":
                return "Nothing happens."
            elif action_type == "look":
                return "Nothing happens."
            elif action_type == "inventory":
                if belief["inventory"]:
                    items = ", ".join(sorted(belief["inventory"]))
                    return f"You are carrying: {items}."
                else:
                    return "You are not carrying anything."
        
        return "Nothing happens."

    def extract_valid_action_forms(self) -> dict[str, list[str]]:
        """Return valid action templates"""
        return self.valid_actions
\end{lstlisting}

\clearpage
\begin{lstlisting}[
  style=worldmodelcode,
  caption={\textbf{BabyAI induced world model.} Egocentric scene representation
  with relative object positions, orientation, and pickup/drop transitions.},
  label={fig:wm-babyai},
]
from abductworld.worldmodel_base import BaseWorldModel
import re
from collections import defaultdict

class BabyaiWorldModel(BaseWorldModel):
    def parse_observation(self, obs_text: str) -> dict:
        """Parse observation text into structured data"""
        result = {
            'objects': [],
            'carrying': None,
            'facing': None,
            'goal': None,
            'walls': True
        }
        
        # Extract goal if present
        goal_match = re.search(r'Your goal: (.+)', obs_text)
        if goal_match:
            result['goal'] = goal_match.group(1)
        
        # Extract carrying information
        carrying_match = re.search(r'You are carrying (?:a|an|the) ([^.]+)', obs_text)
        if carrying_match:
            result['carrying'] = carrying_match.group(1)
        elif 'You are not carrying anything' in obs_text:
            result['carrying'] = None
            
        # Extract facing information
        facing_match = re.search(r'You are facing (?:a|an|the) ([^.]+)', obs_text)
        if facing_match:
            result['facing'] = facing_match.group(1)
        elif (facing_wall := re.search(r'You are facing a wall(?: (\d+) steps away)?', obs_text)):
            steps = facing_wall.group(1) if facing_wall.group(1) else "unknown"
            result['facing'] = f"wall {steps} steps away"
        elif 'You are facing a wall' in obs_text:
            result['facing'] = "wall"
            
        # Extract objects - more robust pattern
        if "In front of you in this room, you can see several objects:" in obs_text:
            object_section = obs_text.split("In front of you in this room, you can see several objects:")[1]
            if "You are facing" in object_section:
                object_section = object_section.split("You are facing")[0]
            
            # Handle the case where objects are listed
            object_lines = object_section.strip()
            
            # Pattern to match objects like: "There is a red box 1 1 steps in front of you and 2 steps to your left."
            object_pattern = r'There is (?:a|an) ([^,]+?) (\d+)(?: right in front of you )?(?:(\d+) steps away|(\d+) steps in front of you)?(?: and (\d+) steps to your (left|right))?[,.]'
            
            for match in re.finditer(object_pattern, object_lines):
                obj_name = match.group(1).strip()
                obj_id = match.group(2)
                
                # Determine front steps
                front_steps = "0"
                if match.group(3):  # steps away
                    front_steps = match.group(3)
                elif match.group(4):  # steps in front of you
                    front_steps = match.group(4)
                elif 'right in front of you' in match.group(0):
                    front_steps = "0"
                
                # Determine lateral position
                lateral_steps = 0
                lateral_dir = 'center'
                if match.group(5) and match.group(6):  # Has lateral position
                    lateral_steps = int(match.group(5))
                    lateral_dir = match.group(6)
                
                full_name = f"{obj_name} {obj_id}"
                result['objects'].append({
                    'name': obj_name,
                    'id': obj_id,
                    'full_name': full_name,
                    'front_steps': int(front_steps),
                    'lateral_steps': lateral_steps,
                    'lateral_dir': lateral_dir
                })
        
        return result

    def init_belief(self):
        """Initialize belief state with latent support"""
        return {
            'objects': [],
            'carrying': None,
            'facing': None,
            'position': (0, 0),  # (front, lateral) coordinates
            'orientation': 0,    # 0=north, 1=east, 2=south, 3=west
            'goal': None,
            'latent_variables': {},  # Add latent support
            'facts': set(),         # Track known facts
            'hypotheses': {},       # Track hypotheses about hidden state
            'frontier': set(),      # Track frontier of exploration
            'hidden_state': {}      # General hidden state tracking
        }

    def correct_belief(self, belief_prior, obs_text: str):
        """Correct belief state based on observation"""
        parsed = self.parse_observation(obs_text)
        belief = belief_prior.copy()
        
        # Deep copy lists and dicts
        belief['objects'] = [obj.copy() for obj in parsed['objects']]
        belief['carrying'] = parsed['carrying']
        belief['facing'] = parsed['facing']
        belief['goal'] = parsed['goal']
        
        # Maintain latent variables
        if 'latent_variables' not in belief:
            belief['latent_variables'] = {}
        if 'facts' not in belief:
            belief['facts'] = set()
        if 'hypotheses' not in belief:
            belief['hypotheses'] = {}
        if 'frontier' not in belief:
            belief['frontier'] = set()
        if 'hidden_state' not in belief:
            belief['hidden_state'] = {}
        
        return belief

    def predict_belief(self, belief, action: str):
        """Predict next belief state given action"""
        new_belief = {}
        for k, v in belief.items():
            if isinstance(v, list):
                new_belief[k] = v.copy()
            elif isinstance(v, set):
                new_belief[k] = v.copy()
            elif isinstance(v, dict):
                new_belief[k] = v.copy()
            else:
                new_belief[k] = v
        
        action_lower = action.lower().strip()
        
        if action_lower.startswith('turn left'):
            new_belief['orientation'] = (new_belief['orientation'] - 1) % 4
        elif action_lower.startswith('turn right'):
            new_belief['orientation'] = (new_belief['orientation'] + 1) % 4
        elif action_lower.startswith('move forward'):
            # Movement doesn't change object positions in this environment
            pass
        elif action_lower.startswith('pickup'):
            # Remove picked up object from scene
            obj_name_match = re.search(r'pickup ([^,]+)$', action_lower)
            if obj_name_match:
                obj_name = obj_name_match.group(1)
                new_belief['objects'] = [obj for obj in new_belief['objects'] if obj['full_name'] != obj_name]
                new_belief['carrying'] = obj_name
        elif action_lower.startswith('drop'):
            # Drop currently carried object in front of agent
            if new_belief['carrying']:
                # Check if there's already an object right in front
                front_objects = [obj for obj in new_belief['objects'] if obj['front_steps'] == 0 and obj['lateral_steps'] == 0]
                if not front_objects:  # Only drop if no object is already in front
                    # Add the dropped object to the front of the agent
                    dropped_obj = {
                        'name': new_belief['carrying'].split(' ')[0],  # Extract object type
                        'id': new_belief['carrying'].split(' ')[1] if len(new_belief['carrying'].split(' ')) > 1 else '1',
                        'full_name': new_belief['carrying'],
                        'front_steps': 0,  # Dropped right in front
                        'lateral_steps': 0,
                        'lateral_dir': 'center'
                    }
                    new_belief['objects'].append(dropped_obj)
                new_belief['carrying'] = None
        elif action_lower.startswith('go to'):
            # Navigation action - moves agent to object location
            obj_name_match = re.search(r'go to ([^,]+)$', action_lower)
            if obj_name_match:
                target_obj_name = obj_name_match.group(1)
                # Find target object
                target_obj = None
                for obj in new_belief['objects']:
                    if obj['full_name'] == target_obj_name:
                        target_obj = obj
                        break
                if target_obj:
                    # Agent moves to object position - object is now in front
                    new_belief['objects'] = [obj for obj in new_belief['objects'] if obj['full_name'] != target_obj_name]
                    target_obj['front_steps'] = 0
                    target_obj['lateral_steps'] = 0
                    target_obj['lateral_dir'] = 'center'
                    new_belief['objects'].append(target_obj)
        elif action_lower.startswith('toggle and go through') or action_lower.startswith('go through'):
            # Door traversal - removes door from scene
            door_match = re.search(r'(?:toggle and go through|go through) ([^,]+)$', action_lower)
            if door_match:
                door_name = door_match.group(1)
                new_belief['objects'] = [obj for obj in new_belief['objects'] if obj['full_name'] != door_name]
        
        return new_belief

    def readout_observation(self, belief, action: str = "") -> str:
        """Generate observation text from belief state"""
        # Handle check available actions specially
        if action.lower().strip() == "check available actions":
            return self._generate_available_actions_response(belief)
        
        lines = []
        
        if belief['goal']:
            lines.append(f"Your goal: {belief['goal']}")
            
        if belief['objects']:
            lines.append("In front of you in this room, you can see several objects:")
            for obj in belief['objects']:
                if obj['lateral_steps'] == 0 and (obj['lateral_dir'] == 'center' or obj['lateral_dir'] == ''):
                    if obj['front_steps'] == 0:
                        lines.append(f"There is a {obj['name']} {obj['id']} right in front of you.")
                    else:
                        lines.append(f"There is a {obj['name']} {obj['id']} {obj['front_steps']} steps in front of you.")
                else:
                    direction = obj['lateral_dir']
                    lines.append(f"There is a {obj['name']} {obj['id']} {obj['front_steps']} steps in front of you and {obj['lateral_steps']} steps to your {direction}.")
        else:
            lines.append("In front of you in this room, you can see several objects: The room has walls around you.")
            
        if belief['facing']:
            lines.append(f"You are facing {belief['facing']}.")
        else:
            lines.append("You are facing a wall.")
            
        if belief['carrying']:
            lines.append(f"You are carrying {belief['carrying']}.")
        else:
            lines.append("You are not carrying anything.")
            
        return " ".join(lines)

    def _generate_available_actions_response(self, belief) -> str:
        """Generate response for 'check available actions' command"""
        # This would normally extract available actions from the belief state
        # For now, we'll return a placeholder that indicates the action was processed
        return "You can take the following actions: turn left, turn right, move forward, check available actions"

    def extract_valid_action_forms(self) -> dict[str, list[str]]:
        """Extract valid action templates"""
        return {
            "go to": ["go to <object>"],
            "pickup": ["pickup <object>"],
            "drop": ["drop"],
            "toggle": ["toggle", "toggle and go through <door>"],
            "go through": ["go through <door>"],
            "move": ["move forward"],
            "turn left": ["turn left"],
            "turn right": ["turn right"],
            "check": ["check available actions"]
        }
\end{lstlisting}

\clearpage
\begin{lstlisting}[
  style=worldmodelcode,
  caption={\textbf{LMRL Gym Maze induced world model.} Compact coordinate-state
  program with local wall constraints and deterministic movement updates.},
  label={fig:wm-maze},
]
from abductworld.worldmodel_base import BaseWorldModel
import re

class MazeWorldModel(BaseWorldModel):
    def parse_observation(self, obs_text: str) -> dict:
        """Parse observation text into structured data."""
        # Extract goal position
        goal_match = re.search(r"The goal is at position (\d+), (\d+)", obs_text)
        if not goal_match:
            return None
        goal_x, goal_y = int(goal_match.group(1)), int(goal_match.group(2))
        
        # Extract current position
        pos_match = re.search(r"Your current position is at position (\d+), (\d+)", obs_text)
        if not pos_match:
            return None
        pos_x, pos_y = int(pos_match.group(1)), int(pos_match.group(2))
        
        # Extract wall information
        walls = []
        if "wall above" in obs_text or "walls above" in obs_text or "above you" in obs_text:
            walls.append("up")
        if "wall below" in obs_text or "walls below" in obs_text or "below you" in obs_text:
            walls.append("down")
        if "wall to your left" in obs_text or "walls to your left" in obs_text or "to your left" in obs_text:
            walls.append("left")
        if "wall to your right" in obs_text or "walls to your right" in obs_text or "to your right" in obs_text:
            walls.append("right")
            
        return {
            "goal": (goal_x, goal_y),
            "position": (pos_x, pos_y),
            "walls": walls
        }

    def init_belief(self):
        """Initialize belief state."""
        return {
            "goal": None,
            "position": None,
            "walls": []
        }

    def correct_belief(self, belief_prior, obs_text: str):
        """Update belief state with new observation."""
        parsed = self.parse_observation(obs_text)
        if parsed is None:
            return belief_prior
            
        # Update belief with parsed information
        belief = belief_prior.copy()
        if parsed["goal"] is not None:
            belief["goal"] = parsed["goal"]
        if parsed["position"] is not None:
            belief["position"] = parsed["position"]
        if parsed["walls"] is not None:
            belief["walls"] = parsed["walls"]
            
        return belief

    def predict_belief(self, belief, action: str):
        """Predict next belief state given action."""
        # Create a copy of current belief
        next_belief = belief.copy()
        
        # Parse action
        action = action.strip().lower()
        
        # Update position based on action if it's a valid move
        if next_belief["position"] is not None:
            x, y = next_belief["position"]
            
            # Apply movement (based on examples: right increases y, down increases x)
            # Fix: when moving down, x should increase; when moving up, x should decrease
            if action == "move up" and "up" not in next_belief["walls"]:
                x -= 1
            elif action == "move down" and "down" not in next_belief["walls"]:
                x += 1
            elif action == "move left" and "left" not in next_belief["walls"]:
                y -= 1
            elif action == "move right" and "right" not in next_belief["walls"]:
                y += 1
                
            next_belief["position"] = (x, y)
            
        return next_belief

    def readout_observation(self, belief, action: str = "") -> str:
        """Convert belief state back to observation text."""
        if belief["goal"] is None or belief["position"] is None:
            return ""
            
        goal_x, goal_y = belief["goal"]
        pos_x, pos_y = belief["position"]
        
        # Start building the observation text
        obs_text = f"The goal is at position {goal_x}, {goal_y}. Your current position is at position {pos_x}, {pos_y}."
        
        # Add wall information
        walls = belief["walls"]
        if walls:
            wall_descriptions = []
            # Maintain deterministic ordering
            if "up" in walls:
                wall_descriptions.append("above you")
            if "down" in walls:
                wall_descriptions.append("below you")
            if "left" in walls:
                wall_descriptions.append("to your left")
            if "right" in walls:
                wall_descriptions.append("to your right")
                
            if len(wall_descriptions) == 1:
                obs_text += f" There is a wall {wall_descriptions[0]}."
            else:
                obs_text += f" There are walls {', '.join(wall_descriptions)}."
        else:
            obs_text += "."
                
        return obs_text

    def extract_valid_action_forms(self) -> dict[str, list[str]]:
        """Return valid action forms for this domain."""
        return {
            "move <X>": ["move up", "move down", "move left", "move right"]
        }
\end{lstlisting}

\clearpage
\begin{lstlisting}[
  style=worldmodelcode,
  caption={\textbf{SciWorld induced world model.} Room-and-object workspace with
  inventory, container contents, and object-state transitions.},
  label={fig:wm-sciworld},
]
from abductworld.worldmodel_base import BaseWorldModel
import re

class SciworldWorldModel(BaseWorldModel):
    def parse_observation(self, obs_text: str) -> dict:
        """
        Parse observation text into structured data.
        """
        # Basic parsing - extract key elements
        parsed = {
            "raw_text": obs_text,
            "room": "",
            "objects": [],
            "inventory": [],
            "actions": [],
            "task": "",
            "room_contents": {},
            "object_states": {},
            "container_contents": {}
        }
        
        # Extract room name
        room_match = re.search(r"This room is called the ([^.]+)\.", obs_text)
        if room_match:
            parsed["room"] = room_match.group(1).strip()
            
        # Extract task description
        task_match = re.search(r"Task description:\s*([^.]+(?:\.[^.]+)*)", obs_text)
        if task_match:
            parsed["task"] = task_match.group(1)
            
        # Extract inventory
        inv_match = re.search(r"In your inventory, you see:\s*((?:\s*.+)+?)(?=\n\n|\Z)", obs_text)
        if inv_match:
            items = inv_match.group(1).strip().split("\n")
            parsed["inventory"] = [item.strip() for item in items if item.strip() and item.strip() != "nothing"]
        elif "In your inventory, you see:" in obs_text and "nothing" in obs_text:
            parsed["inventory"] = []
            
        # Extract room contents
        contents_match = re.search(r"In it, you see:((?:\n\t.+(?:\n\t\t.+)*)+)", obs_text)
        if contents_match:
            contents_text = contents_match.group(1)
            # Parse objects and their states/contents
            lines = contents_text.strip().split('\n')
            current_parent = None
            for line in lines:
                if line.startswith('\t\t'):
                    # This is content of a container
                    if current_parent:
                        content = line.strip()
                        if current_parent not in parsed["container_contents"]:
                            parsed["container_contents"][current_parent] = []
                        parsed["container_contents"][current_parent].append(content)
                elif line.startswith('\t'):
                    # This is an object
                    obj_desc = line.strip()
                    obj_name = obj_desc.split('.')[0] if '.' in obj_desc else obj_desc
                    parsed["objects"].append(obj_name)
                    current_parent = obj_name
                    
                    # Check for states like open/closed, on/off
                    if 'open' in obj_desc:
                        parsed["object_states"][obj_name] = 'open'
                    elif 'closed' in obj_desc:
                        parsed["object_states"][obj_name] = 'closed'
                    elif 'on' in obj_desc:
                        parsed["object_states"][obj_name] = 'on'
                    elif 'off' in obj_desc:
                        parsed["object_states"][obj_name] = 'off'
                        
        return parsed

    def init_belief(self):
        """
        Initialize belief state with latent support.
        """
        return {
            "room": "unknown",
            "objects": {},
            "inventory": [],
            "task": "",
            "room_contents": {},
            "object_states": {},
            "container_contents": {},
            "latent_variables": {
                "door_states": {},
                "container_states": {},
                "object_locations": {}
            },
            "facts": set(),
            "hypotheses": [],
            "frontier": set(),
            "hidden_state": {}
        }

    def correct_belief(self, belief_prior, obs_text: str):
        """
        Update belief state based on new observation.
        """
        parsed = self.parse_observation(obs_text)
        belief = belief_prior.copy()
        
        if parsed["room"]:
            # Fix room name normalization issue
            room_name = parsed["room"].strip()
            if room_name.startswith("the "):
                room_name = room_name[4:]
            elif room_name.startswith("LOC "):
                room_name = room_name[4:]
            belief["room"] = room_name
            
        if parsed["task"]:
            belief["task"] = parsed["task"]
            
        if parsed["inventory"]:
            belief["inventory"] = parsed["inventory"].copy()
        elif "In your inventory, you see:" in obs_text and "nothing" in obs_text:
            belief["inventory"] = []
        # Only clear inventory if explicitly stated as empty, otherwise preserve it
            
        # Update room contents
        if parsed["objects"]:
            belief["room_contents"] = {obj: True for obj in parsed["objects"]}
            
        # Update object states
        belief["object_states"].update(parsed["object_states"])
        
        # Update container contents
        belief["container_contents"].update(parsed["container_contents"])
        
        # Maintain latent variables
        if "latent_variables" not in belief:
            belief["latent_variables"] = {
                "door_states": {},
                "container_states": {},
                "object_locations": {}
            }
            
        if "facts" not in belief:
            belief["facts"] = set()
            
        if "hypotheses" not in belief:
            belief["hypotheses"] = []
            
        if "frontier" not in belief:
            belief["frontier"] = set()
            
        if "hidden_state" not in belief:
            belief["hidden_state"] = {}
            
        return belief

    def predict_belief(self, belief, action: str):
        """
        Predict next belief state given current belief and action.
        """
        belief_next = belief.copy()
        
        # Handle specific actions that change state
        if action.startswith("open "):
            obj_name = action[5:]  # Remove "open "
            if obj_name not in belief_next["object_states"]:
                belief_next["object_states"][obj_name] = "open"
            else:
                belief_next["object_states"][obj_name] = "open"
                
        elif action.startswith("pick up "):
            obj_name = action[8:]  # Remove "pick up "
            if obj_name not in belief_next["inventory"]:
                belief_next["inventory"].append(obj_name)
            # Remove from room contents if present
            if obj_name in belief_next.get("room_contents", {}):
                del belief_next["room_contents"][obj_name]
                
        elif action.startswith("put down "):
            obj_name = action[9:]  # Remove "put down "
            if obj_name in belief_next["inventory"]:
                belief_next["inventory"].remove(obj_name)
                
        elif action.startswith("go to "):
            room_name = action[6:]  # Remove "go to "
            # Fix room name normalization
            if room_name.startswith("the "):
                room_name = room_name[4:]
            elif room_name.startswith("LOC "):
                room_name = room_name[4:]
            belief_next["room"] = room_name
            
        elif action.startswith("examine ") or action.startswith("look at "):
            # These actions can change the current room context
            location = action[8:] if action.startswith("examine ") else action[8:]  # Remove "examine " or "look at "
            # If it's a room name, update the current room
            rooms = ["kitchen", "hallway", "greenhouse", "bathroom", "living room", "bedroom", "workshop", "art studio", "foundry"]
            # Normalize location name
            loc_clean = location
            if loc_clean.startswith("the "):
                loc_clean = loc_clean[4:]
            if loc_clean in rooms:
                belief_next["room"] = loc_clean
                
        # Ensure all required latent fields exist
        required_fields = ["latent_variables", "facts", "hypotheses", "frontier", "hidden_state"]
        for field in required_fields:
            if field not in belief_next:
                if field == "latent_variables":
                    belief_next[field] = {
                        "door_states": {},
                        "container_states": {},
                        "object_locations": {}
                    }
                elif field == "facts":
                    belief_next[field] = set()
                elif field == "hypotheses":
                    belief_next[field] = []
                elif field == "frontier":
                    belief_next[field] = set()
                elif field == "hidden_state":
                    belief_next[field] = {}
                    
        return belief_next

    def readout_observation(self, belief, action: str = "") -> str:
        """
        Generate observation text from belief state.
        """
        if action.startswith("look around") or action == "look around":
            # Generate detailed room description
            room_name = belief.get('room', 'unknown')
            obs_lines = [f"This room is called the {room_name}."]
            
            if belief.get("room_contents") or belief.get("inventory"):
                obs_lines.append("In it, you see:")
                
                # Add room objects
                for obj_name in belief.get("room_contents", {}):
                    obj_line = f"\t{obj_name}"
                    if obj_name in belief.get("object_states", {}):
                        state = belief["object_states"][obj_name]
                        if state in ["open", "closed"]:
                            obj_line += f". The {obj_name} is {state}."
                        elif state in ["on", "off"]:
                            obj_line += f", which is turned {state}."
                    
                    # Check for container contents
                    if obj_name in belief.get("container_contents", {}):
                        contents = belief["container_contents"][obj_name]
                        if contents:
                            obj_line += f". On the {obj_name} is: {', '.join(contents)}."
                        else:
                            obj_line += f". On the {obj_name} is: nothing."
                    elif any(keyword in obj_name for keyword in ["drawer", "cupboard", "fridge", "oven", "freezer"]):
                        obj_line += ". The door is closed."
                        
                    obs_lines.append(obj_line)
                
                # Always add the agent to room contents when looking around
                obs_lines.append("\tthe agent")
                    
            return "\n".join(obs_lines)
            
        elif action.startswith("inventory") or action == "inventory":
            if belief.get("inventory"):
                items = "\n\t".join(belief["inventory"])
                return f"In your inventory, you see:\n\t{items}"
            else:
                return "In your inventory, you see:\n\tnothing"
                
        elif action.startswith("task") or action == "task":
            if belief.get("task"):
                return f"Task description:\n{belief['task']}"
            else:
                return "No task specified."
                
        elif action.startswith("open "):
            obj_name = action[5:]  # Remove "open "
            return f"The {obj_name} is now open."
            
        elif action.startswith("look in "):
            container_name = action[8:]  # Remove "look in "
            if container_name in belief.get("container_contents", {}):
                contents = belief["container_contents"][container_name]
                if contents:
                    content_list = "\n\t".join(contents)
                    return f"Inside the {container_name} is: \n\t{content_list}"
                else:
                    return f"Inside the {container_name} is: \n\tnothing"
            else:
                return f"Inside the {container_name} is: \n\tnothing"
                
        elif action.startswith("pick up "):
            obj_name = action[8:]  # Remove "pick up "
            return f"You move the {obj_name} to the inventory."
            
        elif action.startswith("focus on "):
            obj_name = action[9:]  # Remove "focus on "
            return f"You focus on the {obj_name}."
            
        elif action.startswith("look at "):
            obj_name = action[8:]  # Remove "look at "
            # Try to find object description in belief
            if obj_name in belief.get("object_states", {}):
                state = belief["object_states"][obj_name]
                if "door" in obj_name:
                    return f"A door to the {obj_name.replace(' door', '')} (that is {state})"
                else:
                    return f"a {obj_name}, which is turned {state}."
            elif obj_name in ["kitchen", "hallway", "greenhouse", "bathroom", "living room", "bedroom", "workshop", "art studio", "foundry"]:
                # Looking at a room
                return f"This room is called the {obj_name}."
            else:
                return f"a {obj_name}"
                
        elif action.startswith("examine "):
            obj_name = action[8:]  # Remove "examine "
            if obj_name in ["kitchen", "hallway", "greenhouse", "bathroom", "living room", "bedroom", "workshop", "art studio", "foundry"]:
                # Examining a room - return basic room description
                return f"This room is called the {obj_name}."
            else:
                return f"a {obj_name}"
                
        elif action.startswith("go to "):
            room_name = action[6:]  # Remove "go to "
            # Normalize room name
            if room_name.startswith("the "):
                room_name = room_name[4:]
            return f"You move to the {room_name}."
            
        elif action == "wait" or action == "wait1":
            return "You decide to wait for 10 iterations."
            
        elif action == "":
            # Default response for ambiguous situations
            return "Ambiguous request: Please enter the number for the action you intended (or blank to cancel):"
            
        else:
            # For unhandled actions, return a generic but non-empty response
            return f"You perform the action: {action}"

    def extract_valid_action_forms(self) -> dict[str, list[str]]:
        """
        Return valid action templates.
        """
        return {
            "look at": ["look at <object>"],
            "examine": ["examine <object>"],
            "focus on": ["focus on <object>"],
            "pick up": ["pick up <object>"],
            "put down": ["put down <object>"],
            "move": ["move <object> to <location>"],
            "go to": ["go to <location>"],
            "open": ["open <object>"],
            "close": ["close <object>"],
            "activate": ["activate <object>"],
            "deactivate": ["deactivate <object>"],
            "connect": ["connect <object> to <object>"],
            "disconnect": ["disconnect <object>"],
            "use": ["use <object> on <object>"],
            "pour": ["pour <object> in <object>"],
            "dunk": ["dunk <object> in <object>"],
            "mix": ["mix <object>"],
            "read": ["read <object>"],
            "inventory": ["inventory"],
            "task": ["task"],
            "look around": ["look around"],
            "wait": ["wait"],
            "wait1": ["wait1"],
            "look in": ["look in <object>"]
        }
\end{lstlisting}

\clearpage
\lstinputlisting[
  style=worldmodelcode,
  caption={\textbf{TextCraft induced world model.} Resource-graph belief over
  inventory counts, crafting recipes, and obtainability constraints.},
  label={fig:wm-textcraft},
]{generated_world_models/textcraft_benchmark_world_model.py}

\clearpage
\lstinputlisting[
  style=worldmodelcode,
  caption={\textbf{WebShop induced world model.} Browser-state machine over page
  type, search results, product options, and purchase completion.},
  label={fig:wm-webshop},
]{generated_world_models/webshop_benchmark_world_model.py}

\clearpage
\lstinputlisting[
  style=worldmodelcode,
  caption={\textbf{Wordle induced world model.} Constraint-solving belief over
  candidate target words with guess/feedback history and green/yellow/black
  consistency filtering (no hardcoded target).},
  label={fig:wm-wordle},
]{generated_world_models/wordle_benchmark_world_model.py}

\clearpage
\twocolumn

\end{document}